%% file: main.tex
\documentclass{article}

\PassOptionsToPackage{numbers,compress}{natbib}
\PassOptionsToPackage{table,dvipsnames}{xcolor}
\usepackage[main, final]{neurips_2026}  %

\usepackage[utf8]{inputenc}
\usepackage[T1]{fontenc}
\usepackage{url}
\usepackage{amsfonts}
\usepackage{amsmath}
\usepackage{nicefrac}
\usepackage{microtype}
\usepackage{graphicx}
\usepackage{booktabs}
\usepackage{multirow}
\usepackage{array}
\usepackage{xcolor}
\usepackage{pifont}
\usepackage{wrapfig}
\usepackage{xspace}
\usepackage[accsupp]{axessibility}
\usepackage{hyperref}

\makeatletter
\DeclareRobustCommand\onedot{\futurelet\@let@token\@onedot}
\def\@onedot{\ifx\@let@token.\else.\null\fi\xspace}
\makeatother

\newcommand{\best}[1]{\textcolor{red}{\textbf{#1}}}
\newcommand{\second}[1]{\textcolor{blue}{#1}}
\newcommand{\Yes}{\textcolor{ForestGreen}{\ding{51}}}
\newcommand{\No}{\textcolor{BurntOrange}{\ding{55}}}

\title{ODDR: One-Step Deshadow Diffusion via \\ Reward Guidance}

\author{%
  Junseong Shin$^{1}$ \quad Kijun Kim$^{2}$ \quad Minseong Kim$^{2}$ \quad Dongjin Kim$^{3}$ \quad Tae Hyun Kim$^{1,3\dagger}$ \\[0.5em]
  $^{1}$Department of Artificial Intelligence, Hanyang University \\
  $^{2}$Department of Applied Artificial Intelligence, Hanyang University \\
  $^{3}$Department of Computer Science, Hanyang University \\
  \texttt{\{junsung6140, kijunkim, idecotu, dongjinkim, taehyunkim\}@hanyang.ac.kr} \\[0.3em]
  {\small $^{\dagger}$Corresponding author}
}
\begin{document}

\maketitle

\begin{abstract}
Recent advances in deep learning for shadow removal have significantly enhanced image quality and realism.
However, most approaches rely on real-world paired datasets, which are costly to collect and often limited in scene diversity, leading to limited generalization. 
To address these limitations, we propose One-step Deshadow Diffusion via Reward guidance (ODDR), a new framework that achieves efficient and high-fidelity shadow removal without relying on real-world paired supervision.
Our method begins with One-step Deshadow Diffusion (ODD), a baseline model trained on synthetic shadow data for efficient one-step shadow-free reconstruction.
We further adapt ODD into ODDR using ShadowReward. In contrast to traditional, annotation-heavy approaches, ShadowReward is the first reward model for shadow removal trained entirely without human annotation.
It learns to mimic human perceptual judgments by ranking synthetically generated images with controlled degradations, such as texture distortion and boundary artifacts.
This reward-guided fine-tuning enables ODDR to narrow the synthetic-to-real domain gap. 
Extensive experiments show that ODD achieves strong performance without relying on real-world paired supervision,
and ODDR further improves the results, narrowing the gap to fully supervised methods trained on real-world paired data
while maintaining higher computational efficiency as a single-step model.
Our project page is available \href{https://junsung6140.github.io/oddr_project_page/}{here}.
\end{abstract}

\vspace{-1.0em}
\section{Introduction}
\vspace{-0.5em}
Shadows introduce strong illumination mismatches and visible boundary transitions that harm perceptual quality~\cite{zhang2018improving} and hinder downstream vision tasks, such as text recognition~\cite{lin2020bedsr} and semantic segmentation~\cite{jung2009efficient}.
The goal of shadow removal is to eliminate these artifacts and restore images that are both natural in appearance and suitable for subsequent applications like robotics~\cite{saleh2025fieldnet, abdul2024shadow}.
\vspace{-0.25em}

Recent learning-based approaches substantially improve shadow removal quality through CNN and Transformer-based architectures~\cite{cun2020towards, chen2021canet, guo2023shadowformer, xiao2024homoformer}.
However, many of these approaches require strict supervision using paired real-world data, necessitating precise alignment between shadow images and their shadow-free ground truths.
Unfortunately, collecting such real-world pairs is labor-intensive and often unreliable: scene changes, exposure variation, and illumination fluctuations make alignment difficult, and the resulting datasets are typically limited in diversity.
This limitation leads to a noticeable performance gap between controlled benchmarks and real-world scenarios, where shadows exhibit complex geometry and color interactions.
While previous works have proposed alternatives to bypass the need for real-world paired datasets, these approaches often come with their own inherent drawbacks.
GAN-based~\cite{hu2019mask,liu2021shadow,kubiak2024s3r} approaches rely on adversarial objectives to translate shadowed images into shadow-free ones, but often suffer from unstable training and inconsistent restoration quality.
Synthetic dataset generation~\cite{inoue2020learning} provides paired supervision by rendering shadows onto clean images, yet models trained on synthetic pairs often struggle to remove real-world shadows due to domain gaps.
Recent diffusion-based approaches~\cite{guo2023boundary} address perceptual quality using generative priors, but their iterative sampling procedures lead to high inference latency, which limits practical usage in time-sensitive applications.
Therefore, we aim to elevate perceptual fidelity and inference efficiency without relying on real-world paired datasets, achieving this by preserving the strong perceptual priors of diffusion models.

\begin{wrapfigure}{r}{0.5\textwidth}
  \vspace{-1.5em}
  \centering
  \includegraphics[width=0.5\textwidth]{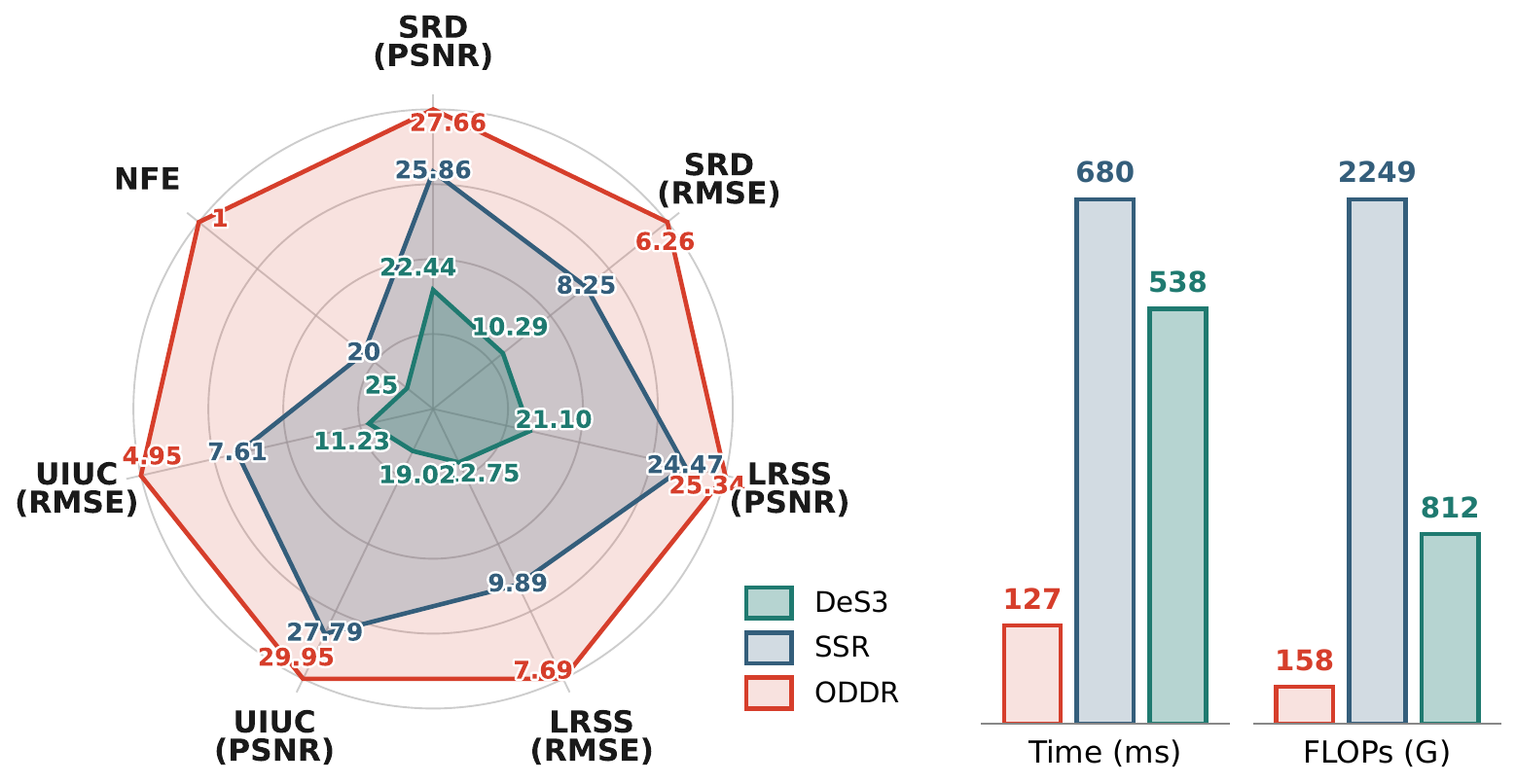}
  \vspace{-2.25 em}
  \caption{We compare our method with diffusion-based shadow removal models, DeS3~\cite{jin2024des3} and SSR~\cite{xu2025detail} on SRD~\cite{qu2017srd}, LRSS~\cite{gryka2015learninglrss}, and UIUC~\cite{guo2012paireduiuc} datasets, and analyze computational efficiency.
  Our ODDR achieves the highest restoration quality and simultaneously demands the least FLOPs, inference time, and NFE (Number of Function Evaluations).}
  \vspace{-2em}
  \label{fig:intro}
\end{wrapfigure}
In this study, we observe that although diffusion priors offer a robust initialization, achieving real-world generalization necessitates training on real shadow artifacts that synthetic datasets fail to capture.
To address this, we introduce a two-stage framework that integrates a diffusion-prior backbone with a novel, shadow-specific perceptual reward model.

In the first stage, we propose a baseline model, \textbf{O}ne-step \textbf{D}eshadow \textbf{D}iffusion (\textbf{ODD}), by adapting a pretrained text-to-image latent diffusion model to perform shadow removal in a single denoising step using only synthetic shadow and shadow-free pairs. 
This diffusion-prior-based training improves structural and color consistency while avoiding common instabilities of GAN-based training. 
In addition, compared to multi-step diffusion~\cite{guo2023shadowdiffusion, guo2023boundary}, this one-step generation drastically reduces the number of function evaluations (NFE) at inference.

In the second stage, we refine ODD into \textbf{O}ne-step \textbf{D}eshadow \textbf{D}iffusion via \textbf{R}eward guidance (\textbf{ODDR}).
Notably, we employ reward-guided optimization directly on images containing real shadows, without requiring corresponding shadow-free references.
Specifically, we introduce \textbf{ShadowReward}, a reward model tailored to shadow removal that approximates human perceptual judgments without relying on manual annotation.
ShadowReward is trained by ranking synthetically corrupted images with controlled degradations that mimic common shadow removal failures, such as texture distortion and boundary artifacts.
During ODDR fine-tuning, the frozen ODD prediction acts as a pseudo target to preserve structural fidelity, while the reward term encourages outputs that receive higher scores from ShadowReward. 
By enabling real-world refinement without real-world paired datasets, this strategy narrows the synthetic-to-real domain gap and enhances perceptual quality without compromising the baseline content or color tone.

Fig.~\ref{fig:intro} summarizes the performance of ODDR compared to other diffusion-based models~\cite{jin2024des3, xu2025detail}, demonstrating its strong efficiency and robust generalization capabilities on unseen datasets~\cite{gryka2015learninglrss, guo2012paireduiuc}.
Through extensive experiments, we show that the proposed method outperforms existing methods that do not use real-world paired data, and substantially narrows the gap to fully supervised methods trained on real-world paired datasets.

\vspace{-1em}
\section{Related Works}
\vspace{-1em}
\subsection{Single-Image Shadow Removal}
\vspace{-0.5em}
Shadow removal methods learn to reconstruct shadow-free images from shadowed inputs by modeling illumination and appearance changes caused by shadows.
Recent approaches utilizing CNNs enhance exposure balance and boundary smoothness via multi-exposure fusion~\cite{fu2021auto}, aggregation based on matting~\cite{cun2020towards}, style harmonization~\cite{wan2022style}, and invertible mappings~\cite{zhu2022bijective}.
Transformer models are particularly effective in capturing global context and long-range dependencies, which facilitates consistent illumination restoration~\cite{guo2023shadowformer,xiao2024homoformer,xu2025omnisr}.
Additionally, diffusion-based approaches contribute to stabilizing texture recovery and preserving details by employing adaptive attention and latent diffusion with detail injection, resulting in more realistic appearances~\cite{guo2023shadowdiffusion, xu2025detail}.
However, diffusion variants, despite their perceptual improvements, come with significant inference costs.
Moreover, these models are still prone to exposure bias in the training data~\cite{wang2018istd, qu2017srd}.

To overcome the reliance on paired datasets, various methods have been extensively explored for shadow removal.
GAN-based approaches employ adversarial or cycle objectives, often with mask guidance, domain-classifier cues, lightness guidance, or non-cyclic unidirectional design~\cite{hu2019mask, jin2021dc, liu2021shadowLG, le2020shadow, liu2021shadow, kubiak2024s3r}, enhancing color fidelity and penumbra smoothness without aligned pairs.
BCDiff~\cite{guo2023boundary}, a diffusion-based method, removes shadows by separating boundary and inner-shadow regions, using non-shadow context for boundary areas and intrinsic decomposition for inner-shadow recovery.
SynShadow~\cite{inoue2020learning} introduces a physics-based synthetic shadow generation method that produces diverse shadow shapes and intensities without paired labels, improving the robustness of deshadowing networks.
However, while such methods eliminate the need for paired data, they typically underperform in shadow removal compared to paired approaches.

\vspace{-0.75em}
\subsection{One-Step Diffusion Models}
\vspace{-0.5em}
Diffusion models~\cite{ho2020ddpm, rombach2022stablediffusion} generate high-quality images through iterative denoising steps that gradually refine random noise into realistic samples. 
While this iterative process enables strong generative capability, it also results in high inference latency and computational overhead in shadow removal~\cite{guo2023shadowdiffusion, xu2025detail}.
To address these issues, recent studies have explored one-step diffusion models~\cite{yin2024one, wu2024osediff} that distill the entire multi-step denoising trajectory into a single forward pass.
This architecture enables efficient task adaptation through lightweight fine-tuning mechanisms, such as LoRA~\cite{hu2022lora}.
These one-step models have demonstrated impressive efficiency and performance across various image restoration tasks such as super-resolution~\cite{sun2025pixel}, JPEG artifact removal~\cite{guo2025compression}, and deblurring~\cite{liu2025one}.
Inspired by these successes and by recent findings that pretrained diffusion models can be efficiently fine-tuned for single-step prediction~\cite{garcia2025finetuning}, we adapt the one-step diffusion paradigm to shadow removal.
Rather than the one-step formulation itself, our contribution lies in the architectural designs that make it effective for shadow removal and in the annotation-free reward guidance described below.

\vspace{-0.75em}
\subsection{Reward Models and Preference Supervision}
\vspace{-0.5em}
Reward modeling has become a central component in reinforcement learning with human feedback, where a learned reward function is optimized to align model outputs with human preferences~\cite{ziegler2019fine, bai2022training}.
Such approaches have recently gained substantial attention in large language models~\cite{ouyang2022training, wu2023fine}.
Beyond language modeling, preference-based learning has also been extended to various computer vision tasks, including face restoration~\cite{wu2025diffusionreward}, image generation~\cite{xu2023imagereward, liang2024rich, kirstain2023pick, wu2023human}, video generation~\cite{li2024t2v}, and 3D generation~\cite{ye2024dreamreward}.
These methods demonstrate that reward supervision can effectively improve perceptual quality by aligning model outputs with human judgments.
However, existing reward models typically rely on preference datasets constructed from large-scale human annotations, which are costly and labor-intensive to collect.
This dependency limits their scalability and practical applicability, particularly in low-level vision tasks where fine-grained perceptual differences are critical.
To alleviate this limitation, recent image restoration studies have explored automated preference supervision using vision-language models (VLMs)~\cite{bai2025qwen2} or perceptual Image Quality Assessment (IQA) metrics in super-resolution tasks~\cite{wu2025dp,qiao2025realsr}.
Nevertheless, such strategies have not yet been specifically tailored to the shadow removal problem.
In this work, we introduce the first reward model designed for shadow removal. 
Notably, our proposed model is trained entirely without human annotations, yet provides human-aligned perceptual guidance to improve restoration quality.

\vspace{-0.5em}
\section{Proposed Method}
\vspace{-1em}
\subsection{Overall Framework}
\begin{figure*}[t]
    \centering
    \includegraphics[width=0.99\linewidth]{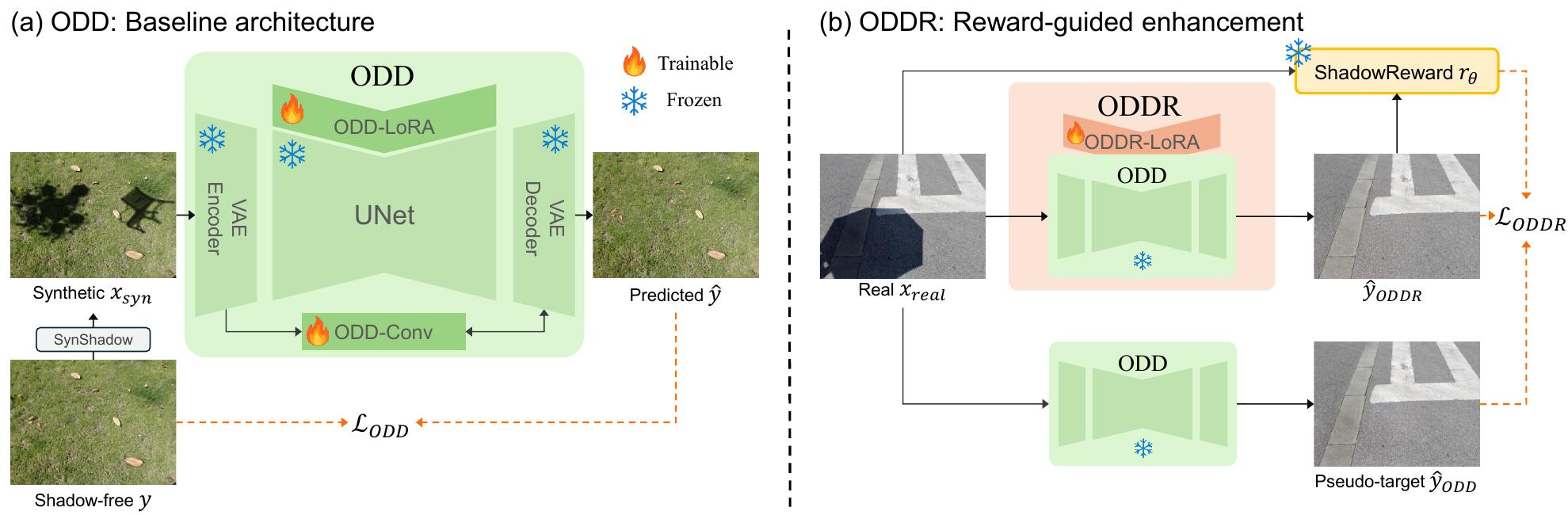}
    \vspace{-1.0em}
    \caption{\textbf{Overall training pipeline.} (a) The baseline ODD is trained on synthetic shadow images ($x_{\text{syn}}$), optimizing the ODD-LoRA and ODD-Conv modules. (b) ODDR is then fine-tuned from ODD parameters using real-world shadow images ($x_{\text{real}}$) guided by ShadowReward ($r_\theta$) model, optimizing ODDR-LoRA module.}
    \label{fig:main}
    \vspace{-1.5em}
\end{figure*}

\vspace{-0.5em}
Our proposed framework consists of two primary components: 
(i) One-step Deshadow Diffusion (ODD), a cost-effective one-step latent diffusion for deshadowing, and (ii) One-step Deshadow Diffusion via Reward guidance (ODDR), an enhanced version of ODD for real-world scenarios guided by a preference-based reward model (ShadowReward).
As shown in Fig.~\ref{fig:main}, we first train ODD by adapting Stable Diffusion (SD)~\cite{rombach2022stablediffusion} using synthetic shadow and shadow-free pairs~\cite{inoue2020learning}. 
Subsequently, we refine ODD through ShadowReward, which provides perceptual feedback to assist in the fine-tuning process.
This results in our final model, ODDR, which significantly enhances shadow removal performance in real scenarios.
In the following, Sec.~\ref{sec:odd} describes the ODD baseline model, Sec.~\ref{sec:oddr} introduces the reward-guided refinement for ODDR including the ShadowReward model and its human-annotation-free training procedure.

\vspace{-0.5em}
\subsection{One-Step Deshadow Diffusion (ODD)}
\label{sec:odd}
\vspace{-0.5em}
\subsubsection{Motivation}
\vspace{-0.5em}
In general, deshadow models trained without real-world paired datasets struggle to match the performance of fully supervised approaches in real scenarios.
To bridge this gap, we harness the image priors of Stable Diffusion (SD)~\cite{rombach2022stablediffusion} to construct our baseline model, ODD, which significantly enhances deshadowing quality without using the real-world paired dataset.
However, SD relies on iterative denoising that requires many inference steps, leading to high computational cost and latency. 
Furthermore, because intermediate diffusion states remain distant from the final image, providing direct pixel-level feedback from the reward model is very challenging.
To overcome these limitations, we adopt the one-step formulation from OSEDiff~\cite{wu2024osediff}, which replaces stochastic multi-step denoising with a single deterministic mapping that directly predicts a shadow-free image.

\noindent\textbf{Architecture}
The architecture of ODD is built upon SD, which consists of a latent variational autoencoder (VAE) and a UNet-based denoising network pre-trained on large-scale datasets~\cite{schuhmann2022laion}, providing a robust foundation from large-scale pre-training.
We introduce several key modifications to adapt this architecture for the one-step shadow removal task.

First, as shown in Fig.~\ref{fig:main} (a), we adapt the SD UNet with LoRA~\cite{hu2022lora}, inserting rank-16 low-rank adapters into its attention and convolution blocks (ODD-LoRA). 
This partial adaptation enables the model to effectively adjust to the shadow removal task while keeping the original SD weights frozen.
Despite this benefit, simply repurposing the SD architecture can lead to a loss of fine details, which are crucial for the shadow removal task.
To address this limitation, we add lightweight two-layer convolutional skip connections (ODD-Conv) at each encoder-decoder level, which assist in maintaining fine details and recovering high-frequency textures.
Furthermore, since our model does not take text prompts as input, we replace the original SD's text-conditioning modules with learnable self-attention layers.

\noindent\textbf{Training Objective}
Given a shadow-free image $y$, we first synthesize a corresponding shadowed image $x_{\text{syn}}$ through SynShadow~\cite{inoue2020learning} and train ODD to predict $\hat{y}$ that matches $y$ by minimizing $\ell_1$-based reconstruction and perceptual losses:
\begin{equation}
\label{eq:loss_odd}
\mathcal{L}_{\text{ODD}} = \| \hat{y} - y \|_1 + \lambda_{\text{per}}\mathcal{L}_{\text{per}}(\hat{y}, y),
\end{equation}
where $\mathcal{L}_{\text{per}}$ is the LPIPS loss~\cite{zhang2018unreasonable} and $\lambda_{\text{per}}$ is a hyperparameter balancing the two losses.
This objective encourages structural fidelity and perceptual realism, serving as the initialization for subsequent reward-guided fine-tuning.

\vspace{-1.0em}
\subsection{One-Step Deshadow Diffusion via Reward Guidance (ODDR)}
\label{sec:oddr}
\vspace{-0.5em}

Although ODD achieves effective reconstruction as a baseline, it still presents two significant drawbacks.
First, because it is trained exclusively on synthetically shadowed images, a noticeable domain gap remains when applied to real-world shadows with complex lighting and color.
Second, while LPIPS enforces feature-level alignment with the ground truth, it frequently overlooks visually salient artifacts inherent to shadow removal, including texture degradation, noise, color deviations, luminance fluctuations, and boundary seams.
To overcome these limitations, we present ODDR, which adapts the baseline model (ODD) with a perceptual reward model (ShadowReward) to better align with human visual judgments.

As illustrated in Fig.~\ref{fig:main}~(b), our process begins with two parallel instances of the pretrained ODD (top and bottom branches) with frozen weights.
Given a real-world shadow input $x_{\text{real}}$, the bottom branch (ODD) generates a pseudo-target $\hat{y}_{\text{ODD}}$.
Subsequently, the top branch (ODDR) undergoes partial fine-tuning through LoRA~\cite{hu2022lora} (ODDR-LoRA), and is optimized by the objective function with a frozen reward function $r_\theta$ as follows:
\begin{equation}
\begin{aligned}
\label{eq:loss_oddr}
\mathcal{L}_{\text{ODDR}} =  & \| \hat{y}_{\text{ODDR}} - \hat{y}_{\text{ODD}} \|_2  + \lambda_{\text{align}} \, \mathcal{L}_{\text{per}}(\hat{y}_{\text{ODDR}}, \hat{y}_{\text{ODD}}) 
- \lambda_r \, r_\theta(x_{\text{real}}, \hat{y}_{\text{ODDR}}),
\end{aligned}
\end{equation}
where $\hat{y}_{\text{ODDR}}$ denotes the ODDR (top branch) output.
To train ODDR without ground-truth shadow-free images, the loss function combines two main objectives.
First, $\ell_2$ loss and perceptual loss $\mathcal{L}_{\text{per}}$ (weighted by $\lambda_{\text{align}}$) maintain structural fidelity to the baseline prediction (ODD).
Second, a frozen reward function $r_\theta$ (weighted by $\lambda_r$) subtracts a reward score that reflects human preference from the overall loss, successfully preserving data fidelity while guiding the model toward perceptually enhanced, boundary-consistent restorations.

\noindent\textbf{Ranking Deshadow Preference (ShadowReward)}
\label{sec:shadowreward}
In this work, we employ a reward model to fine-tune the baseline ODD model, which significantly enhances the perceptual quality of the final output.
In the context of the reward model, human visual preference serves as a natural standard for assessing the quality of shadow removal results, and observers primarily evaluate the quality of restoration through boundary smoothness, color consistency, and texture realism~\cite{vasluianu2024ntirewsrdp}.
However, collecting the human-annotated preference data required to train such reward models is a time-consuming and expensive process.
To address this problem, we present \textbf{ShadowReward}, a reward model designed to mimic human preference without the need for human annotation to assess the quality of shadow removal.

Previous studies have shown that the level of image degradation influences perceptual preference, with more severely degraded images consistently judged as less favorable~\cite{ponomarenko2015image, larson2010most}. 
Motivated by this observation, we design a training strategy that leverages controlled degradations to induce preference ordering, enabling the reward model to learn human-like judgments without explicit annotation.
Specifically, to simulate scenarios where conventional shadow removal models generally do not perform well, we corrupt images with specific degrees of texture distortion, noise, chromatic shift, luminance change, and boundary artifacts.
Subsequently, we train ShadowReward to rank these corrupted images according to their degree of perceptual degradation.

\begin{figure}[t]
    \centering
    \includegraphics[width=0.70\linewidth]
    {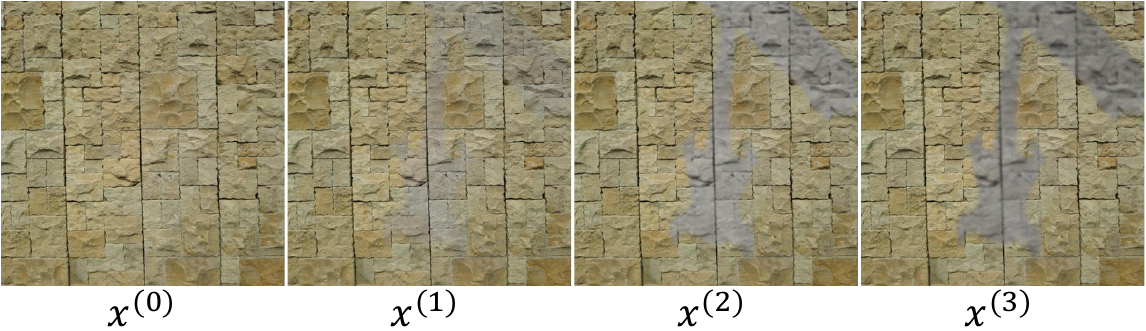}
    \vspace{-1em}
    \caption{An example list $\{x^{(j)}\}_{j=0}^3$ of the human-annotation-free dataset. Larger $j$ means more degradation.}
    \label{fig:human_free_dataset}
    \vspace{-2.0em}
\end{figure}

\noindent\textbf{Human-Annotation-Free Dataset}
To train our ShadowReward model, we construct a Human-Annotation-Free dataset to enable preference learning without human labels.
Specifically, following SynShadow~\cite{inoue2020learning}, we first generate a synthetic shadow image $x_{\text{syn}}$ from a corresponding shadow-free image $y$ and a random shadow mask. %
We then create a list of degraded images $\{x^{(j)}\}_{j=0}^J$, where each $x^{(j)}$ exhibits a progressively increasing degree of degradation corresponding to its index $j$. 
As described in Fig.~\ref{fig:human_free_dataset}, a higher index indicates more degradation and lower perceptual quality, while $x^{(0)}$ is assigned as the clean image $y$. Through empirical evaluation, we set $J=3$.

To manage the extent of each degradation, we initially establish a series of degradation types along with their respective degradation levels.
These include texture distortion, additive noise, chromatic shift, luminance change, and boundary artifacts.
Each is applied with a range of degradation levels to control degradation strength. 
This process allows us to construct diverse and perceptually ordered training samples without any human annotation.
Detailed settings are provided in our Appendix (Sec.~\ref{supp:human_anno_free}).

\begin{wrapfigure}{r}{0.5\linewidth}
\vspace{-2.5em}
\centering
\includegraphics[width=0.9\linewidth]{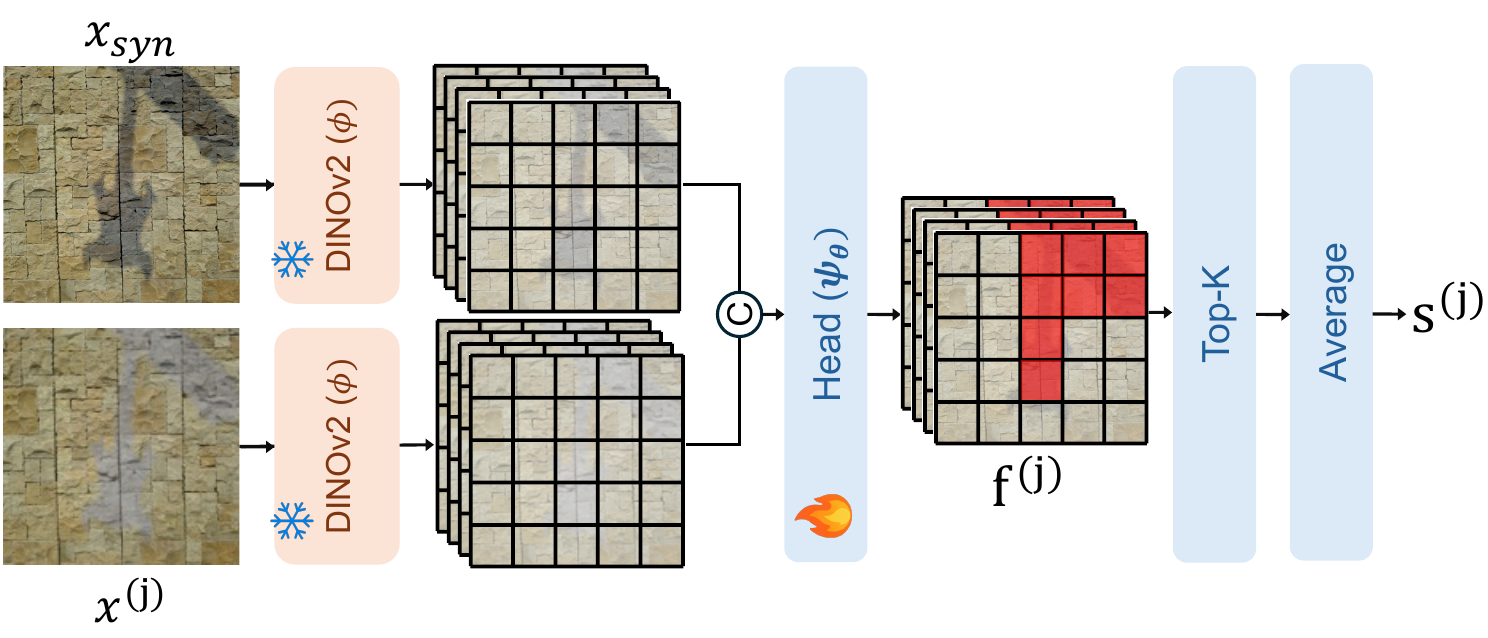}
\vspace{-1em}
\caption{Overview of ShadowReward.}
\label{fig:shadowreward_arch}
\vspace{-1em}
\end{wrapfigure}

\noindent\textbf{ShadowReward Architecture}
In contrast to conventional reward models based on VLMs~\cite{radford2021learning, li2022blip}, our shadow removal task requires nuanced spatial reasoning and sensitivity to local texture/illumination variations rather than high-level semantics.
Thus, as shown in Fig.~\ref{fig:shadowreward_arch}, we use DINOv2~\cite{oquab2023dinov2} in our ShadowReward to capture spatial correspondence and discriminative local features for shadow removal.
Specifically, we extract patch-level tokens from both the shadow input $x_{\text{syn}}$ and a corrupted image $x^{(j)}$ using the DINOv2 encoder ($\phi$). 
Next, these token sequences are concatenated and passed to a lightweight convolutional head $\psi_\theta$, which produces a spatial map as follows:
\begin{equation}
f^{(j)} = \psi_\theta([\phi(x_{\text{syn}}), \phi(x^{(j)})]),
\end{equation}
where $[\cdot,\cdot]$ denotes the concatenation operation.
The concatenated features are processed by $\psi_\theta$ to capture the relative discrepancies between the shadowed and corrupted images, producing patch-wise perceptual features $f^{(j)}$.
Then, $f^{(j)}$ is mapped into a final scalar reward $s^{(j)}$, which serves as a perceptual quality score for $x^{(j)}$.   
To compute this score, we apply Top-$K$ pooling, averaging only the largest
responses of $f^{(j)}$ so that the reward focuses on informative regions that
best reflect perceptual differences. 
In practice, we set $K=0.1$ (top 10\% responses).
The scalar reward is computed as
\begin{equation}
s^{(j)} = \frac{1}{|\mathcal{T}_K|}\sum_{i \in \mathcal{T}_K} f^{(j)}_i,
\end{equation}
where $\mathcal{T}_K$ denotes the indices of the top-$K$ responses in $f^{(j)}$.

\noindent\textbf{Pre-training ShadowReward}
Assessing perceptual quality through ShadowReward entails comparing differently corrupted images and determining their relative quality.
It is understood that determining relative rankings is easier and more dependable than scoring absolute quality. 
This is because measuring absolute quality is challenging due to the variety of images.
To this end, we train ShadowReward $r_\theta$ to rank corrupted images $\{x^{(j)}\}_{j=0}^J$, assigning higher scores to those with better perceptual quality.
To train ShadowReward concerning this ranking behavior, we utilize the ListMLE loss~\cite{xia2008listwise} denoted by $\mathcal{L}_{\text{list}}$. This loss function models the likelihood of achieving the correct sequence of a list of distorted images, and it is defined as follows:
\begin{equation}
\label{eq:loss_list}
\mathcal{L}_{\text{list}}
=
-\frac{1}{J+1}
\sum_{i=0}^{J}
\log
\frac{\exp\!\left(s^{(i)}/\tau\right)}
{\sum_{t=i}^{J}\exp\!\left(s^{(t)}/\tau\right)},
\end{equation}
where the temperature parameter $\tau$ regulates how decisively the model distinguishes between corrupted images in the ranking process.
We set $\tau = 1$ by default.
A higher score $s^{(j)}$ indicates better perceptual quality, meaning that images with fewer artifacts or distortions receive larger rewards.
This loss enforces a higher reward for perceptually better images, which enables the reward model to assess the quality of shadow removal results. 
Note that our ShadowReward model can be trained separately from ODD and ODDR.

\vspace{-1em}
\section{Experiments}
\label{sec:experiments}
\vspace{-0.5em}

\subsection{Experimental Setup}
\vspace{-0.5em}
\noindent\textbf{Datasets}
We use the Adjusted ISTD (AISTD) dataset~\cite{wang2018istd, le2019shadow} for training, which contains 1,870 triplets of shadow, shadow mask, and shadow-free images with a resolution of $480\times640$, where 1,330 triplets are used for training and 540 for testing. 
We utilize different components of the AISTD training set for each stage of our framework.
ODD is trained solely on shadow-free images using synthetic shadow images generated from clean images in this dataset.
For ODDR, we use only shadowed images from the training set.
ShadowReward is trained on our Human-Annotation-Free dataset, which is constructed by augmenting AISTD shadow-free images with random shadow masks with SynShadow~\cite{inoue2020learning}.

To evaluate the generalization ability of our model, we conduct tests on three unseen real-world datasets, SRD~\cite{qu2017srd}, LRSS~\cite{gryka2015learninglrss}, and UIUC~\cite{guo2012paireduiuc}.
The SRD and LRSS datasets contain 408 and 34 test pairs, respectively; LRSS consists of soft shadows, allowing us to assess robustness to faint shadow boundaries.
For the UIUC dataset, we use 44 test pairs with consistent light conditions from the original 76 pairs.
These datasets provide diverse illumination and shadow configurations that differ from AISTD, enabling a fair and rigorous evaluation of cross-dataset generalization.

\begin{table*}[t]
  \centering
  \caption{Quantitative comparison on AISTD~\cite{le2019shadow} dataset.
  \best{Red} and \second{blue} denote the best and second-best results, evaluated among the methods trained without real-world pairs. \Yes~ denotes mask-free methods and~\No~ indicates methods that require shadow masks. $\mathcal{D}$ indicates whether real-world paired data is used for training.}
  \label{tab:shadow_results}
  \resizebox{0.995\linewidth}{!}{
  \begin{tabular}{c|l|c|ccc|ccc|ccc}
    \toprule
    \rowcolor{blue!7}
    \multicolumn{1}{c|}{\multirow{2}{*}{$\mathcal{D}$}} & \multirow{2}{*}{Methods} & \multirow{2}{*}{Mask-free}
      & \multicolumn{3}{c|}{\textbf{Shadow Region}}
      & \multicolumn{3}{c|}{\textbf{Non-Shadow Region}}
      & \multicolumn{3}{c}{\textbf{All Image}} \\
    \rowcolor{blue!7}
     \multicolumn{1}{c|}{\multirow{-2}{*}{$\mathcal{D}$}} & \multirow{-2}{*}{Methods}& \multirow{-2}{*}{Mask-free} & PSNR$\uparrow$ & SSIM$\uparrow$ & RMSE$\downarrow$
      & PSNR$\uparrow$ & SSIM$\uparrow$ & RMSE$\downarrow$
      & PSNR$\uparrow$ & SSIM$\uparrow$ & RMSE$\downarrow$ \\
    \midrule
      & Real-world Shadow Image & -- & 20.83 & 0.927 & 39.01 & 37.46 & 0.985 & 2.40 & 20.46 & 0.894 & 8.40  \\
    \midrule
    \multirow{8}{*}{\rotatebox[origin=c]{90}{w/ real-world pair}}
      & DHAN~\cite{cun2020towards} & \No & 32.92 & 0.988 & 9.6 & 27.15 & 0.971 & 7.4 & 25.66 & 0.956 & 7.8 \\
      & SP+M-Net~\cite{le2019shadow} & \No & 37.60 & 0.990 & 6.3 & 36.02 & 0.976 & 3.0 & 32.94 & 0.962 & 3.5 \\
      & Fu et al.~\cite{fu2021auto} & \No & 36.04 & 0.978 & 6.7 & 31.16 & 0.892 & 3.8 & 29.45 & 0.861 & 4.2 \\
      & BMNet~\cite{zhu2022bijective}& \No & 37.87 & 0.991 & 5.62 & 37.51 & 0.985 & 2.45 & 33.98 & 0.972 & 2.97 \\
      & SG-ShadowNet~\cite{wan2022style}    & \No & 36.80 & 0.990  & 5.93 &  35.57 &  0.978 & 2.92 & 32.46 & 0.962 & 3.41  \\
      & DeS3~\cite{jin2024des3}    & \Yes & 36.49 & 0.989  & 6.59 &  34.72 &  0.972 & 3.32 & 31.39 & 0.957 & 3.86  \\
      & OmniSR~\cite{xu2025omnisr}    & \Yes & 37.01 & 0.990  & 6.57 &  37.82 &  0.984 & 2.40 & 33.35 & 0.970 & 3.08  \\
      & SSR~\cite{xu2025detail}    & \Yes & 39.94 & 0.993  & 4.23 & 38.04  & 0.983  & 2.63 & 35.19 & 0.974 & 2.89 \\
    \midrule
    \multirow{9}{*}{\rotatebox[origin=c]{90}{w/o real-world pair}}
      & MaskShadow-GAN~\cite{hu2019mask}  & \No & 32.19 & 0.984 & 10.8 & 33.44 & 0.974 &  3.8 & 28.81 &  0.946 &  4.8  \\
      & LG-ShadowNet~\cite{liu2021shadowLG}    & \No & 32.45 & 0.982 & 9.65 & 33.73 & 0.975 & 3.25 & 29.22 & 0.947 & 4.30 \\
      & DCS~\cite{jin2021dc}    & \Yes & 32.00 & 0.976 & 10.41 & 33.56 & 0.968 & 3.63 & 28.77 & 0.932 & 4.74 \\
      & Le et al.~\cite{le2020shadow}       & \No & 33.09 & 0.983 & 9.67 & 35.35 & \second{0.978} & 2.82 & 30.15 & 0.951 & 3.94 \\
      & G2R~\cite{liu2021shadow}             & \No & 33.58 & 0.979 &  8.82 & 35.52 &  0.976 &  2.89 & 30.52 &  0.944 &  3.86  \\
      & BCDiff~\cite{guo2023boundary}                   & \No & 35.91 & 0.986 & 7.6 & 37.27 & \best{0.984} & 2.4 & 32.73 & 0.962 & 3.3 \\
      & S3R-Net~\cite{kubiak2024s3r}                   & \Yes & 32.08 & 0.983 & 12.03 & 33.61 & \second{0.978} & 3.55 & 29.05 & 0.952 & 9.94 \\
      & \cellcolor{gray!10}ODD (ours)           & \cellcolor{gray!10}\Yes & \cellcolor{gray!10}\second{36.67} & \cellcolor{gray!10}\second{0.990} & \cellcolor{gray!10}\second{6.27} & \cellcolor{gray!10}\best{37.45} & \cellcolor{gray!10}\best{0.984} & \cellcolor{gray!10}\best{2.49} & \cellcolor{gray!10}\second{33.32} & \cellcolor{gray!10}\second{0.971} & \cellcolor{gray!10}\second{3.11} \\
      & \cellcolor{gray!10}ODDR (ours)      & \cellcolor{gray!10}\Yes & \cellcolor{gray!10}\best{38.37} & \cellcolor{gray!10}\best{0.991} & \cellcolor{gray!10}\best{5.36} & \cellcolor{gray!10}\second{37.34} & \cellcolor{gray!10}\best{0.984} & \cellcolor{gray!10}\second{2.50} & \cellcolor{gray!10}\best{34.20} & \cellcolor{gray!10}\best{0.972} & \cellcolor{gray!10}\best{2.97} \\
    \bottomrule
  \end{tabular}
  }

\end{table*}

\noindent\textbf{Evaluation Metrics}
Following previous works~\cite{le2019shadow, fu2021auto}, 
we use three standard metrics: Root Mean Square Error (RMSE), Peak Signal-to-Noise Ratio (PSNR), and Structural Similarity (SSIM)~\cite{wang2004image}.
RMSE is computed in the LAB color space to measure color consistency, while PSNR and SSIM are calculated in the RGB color space. 
Lower RMSE indicates smaller reconstruction errors, while higher PSNR and SSIM denote better restoration quality.
All metrics are computed on images of size $256\times256$ using the MATLAB evaluation scripts from previous work~\cite{zhu2022efficient}.

\noindent\textbf{Implementation Details}
For ODD, we adopt the pretrained SD v2.1 model~\cite{rombach2022stablediffusion} as our backbone.
The model is trained on $512\times512$ inputs, where smaller images are padded to this size using reflect padding.
For optimization, we use the AdamW optimizer~\cite{loshchilov2017decoupled} with a batch size of 4 for 40K iterations, and a learning rate decaying from $1\times10^{-4}$ to $1\times10^{-6}$ via a cosine annealing scheduler~\cite{loshchilov2016sgdr}.
For ShadowReward, we use $448\times448$ inputs, a batch size of 16, and the AdamW optimizer for 20K iterations with the same learning rate decay strategy as ODD.
For ODDR, we use an identical setup to ODD, except that it is trained for 20K iterations with a fixed learning rate of $1\times10^{-6}$.
All experiments are conducted on a single NVIDIA RTX A6000 GPU.

\begin{figure*}[t]
    \centering
    \vspace{-1em}
    \includegraphics[width=0.995\linewidth]
    {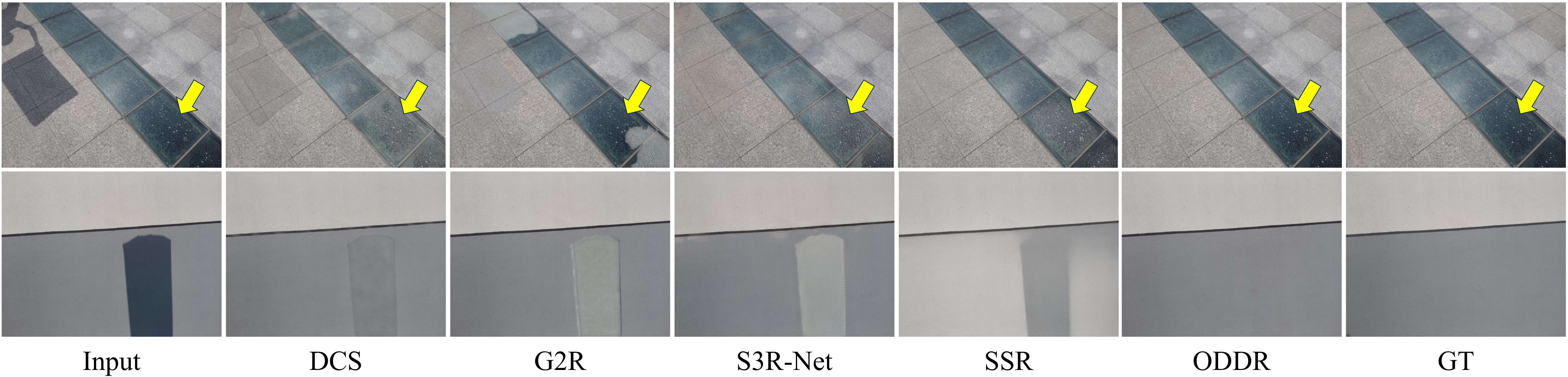}
    \vspace{-2em}
  \caption{Visual comparison on AISTD~\cite{le2019shadow}. Top: non-shadow preservation (yellow arrows mark
  competitor failures). Bottom: shadow removal. Ours performs well in both.}
    \vspace{-1em}
    \label{fig:comparison}
\end{figure*}

\vspace{-1em}
\subsection{Quantitative and Qualitative Comparisons}

We compare the proposed method with various shadow removal approaches across different training data settings. 
For methods trained with real-world pairs, we evaluate against DHAN~\cite{cun2020towards}, SP+M-Net~\cite{le2019shadow}, Fu et al.~\cite{fu2021auto}, BMNet~\cite{zhu2022bijective}, SG-ShadowNet~\cite{wan2022style}, and OmniSR~\cite{xu2025omnisr}. Moreover, we incorporate recent diffusion-based shadow removal models such as DeS3~\cite{jin2024des3} and SSR~\cite{xu2025detail} to evaluate our single-step method over multi-step diffusion processes. 
For methods trained without real-world paired datasets, we compare with MaskShadow-GAN~\cite{hu2019mask}, DCS~\cite{jin2021dc}, LG-ShadowNet~\cite{liu2021shadowLG}, G2R~\cite{liu2021shadow}, BCDiff~\cite{guo2023boundary}, and S3R-Net~\cite{kubiak2024s3r}.

\begin{table*}[t]
\setlength{\tabcolsep}{8pt}
\centering
  \caption{Results on SRD, LRSS, and UIUC. The best results are highlighted in \best{red}.}
\label{tab:lrss_uiuc_results}
\resizebox{0.9\textwidth}{!}{
\begin{tabular}{l|ccc|ccc|ccc}
\toprule
\rowcolor{blue!7}
 & \multicolumn{3}{c|}{\textbf{SRD}~\cite{qu2017srd}} & \multicolumn{3}{c|}{\textbf{LRSS}~\cite{gryka2015learninglrss}} & \multicolumn{3}{c}{\textbf{UIUC}~\cite{guo2012paireduiuc}} \\
\rowcolor{blue!7}
\textbf{Method} & PSNR$\uparrow$ & SSIM$\uparrow$ & RMSE$\downarrow$ & PSNR$\uparrow$ & SSIM$\uparrow$ & RMSE$\downarrow$ & PSNR$\uparrow$ & SSIM$\uparrow$ & RMSE$\downarrow$ \\
\midrule
S3R-Net~\cite{kubiak2024s3r}      & 22.51 & 0.878 & 10.60 & 22.54 & 0.786 & 10.45 & 23.92 & 0.857 &  8.53 \\
DCS~\cite{jin2021dc}              & 21.99 & 0.868 & 11.47 & 22.10 & 0.774 & 11.63 & 23.94 & 0.852 &  9.69 \\
G2R~\cite{liu2021shadow}          & 24.23 & 0.898 &  8.40 & 22.79 & 0.778 &  9.64 & 27.03 & 0.867 &  6.23 \\
DeS3~\cite{jin2024des3}           & 22.44 & 0.900 & 10.29 & 21.10 & 0.767 & 12.75 & 19.02 & 0.822 & 11.23 \\
SSR~\cite{xu2025detail}           & 25.86 & 0.922 &  8.25 & 24.47 & 0.790 &  9.89 & 27.79 & 0.873 &  7.61 \\
\midrule
\rowcolor{gray!10}
ODD                               & 27.38 & 0.941 &  6.46 & 25.12 & 0.797 &  7.84 & 29.25 & 0.886 &  5.10 \\
\rowcolor{gray!10}
ODDR                              & \best{27.66} & \best{0.942} & \best{6.26} & \best{25.34} & \best{0.797} & \best{7.69} & \best{29.95} & \best{0.887} & \best{4.95} \\
\bottomrule
\end{tabular}}
\vspace{-1em}
\end{table*}

\noindent\textbf{Quantitative Results}
Tab.~\ref{tab:shadow_results} reports the main comparison results on the AISTD dataset, with metrics computed on the shadow, non-shadow, and whole-image regions.
First, our ODD performs competitively among methods without using real-world pairs. 
Notably, our final reward-guided model (ODDR) further enhances shadow-region performance, increasing PSNR from 36.67\,dB to 38.37\,dB and reducing RMSE from 6.27 to 5.36.
Since both alignment terms in Eq.~(\ref{eq:loss_oddr}) are minimized at the ODD output, training with $\lambda_r{=}0$ reproduces ODD (Appendix Sec.~\ref{supp:reward_weight}); this gain is therefore attributable to the reward.

While some methods trained on real-world paired datasets show higher PSNR in non-shadow regions than the input, this improvement often arises from illumination discrepancies in paired data, leading models to learn global brightness corrections rather than pure shadow removal~\cite{hu2019mask}.
Therefore, we measure cross-dataset generalization performance on three unseen real-world datasets, SRD, LRSS, and UIUC, without additional training.
For a fair comparison, all results are obtained with the parameters trained on the AISTD dataset.
As shown in Tab.~\ref{tab:lrss_uiuc_results}, our ODD model achieves superior performance compared to conventional approaches, surpassing methods trained both with and without real-world paired data by a significant margin.
Although ODD has strong cross-dataset generalization, ODDR consistently outperforms ODD and achieves higher PSNR and SSIM, and lower RMSE across all datasets, confirming its improved generalization performance.
We note that the pretrained SD prior shared by ODD and ODDR contributes substantially to this generalization, while the reward adds a smaller but consistent refinement on top of it (+0.28 / +0.22 / +0.70\,dB PSNR on SRD / LRSS / UIUC).

\begin{figure}[t]
    \centering
    \includegraphics[width=0.92\linewidth]{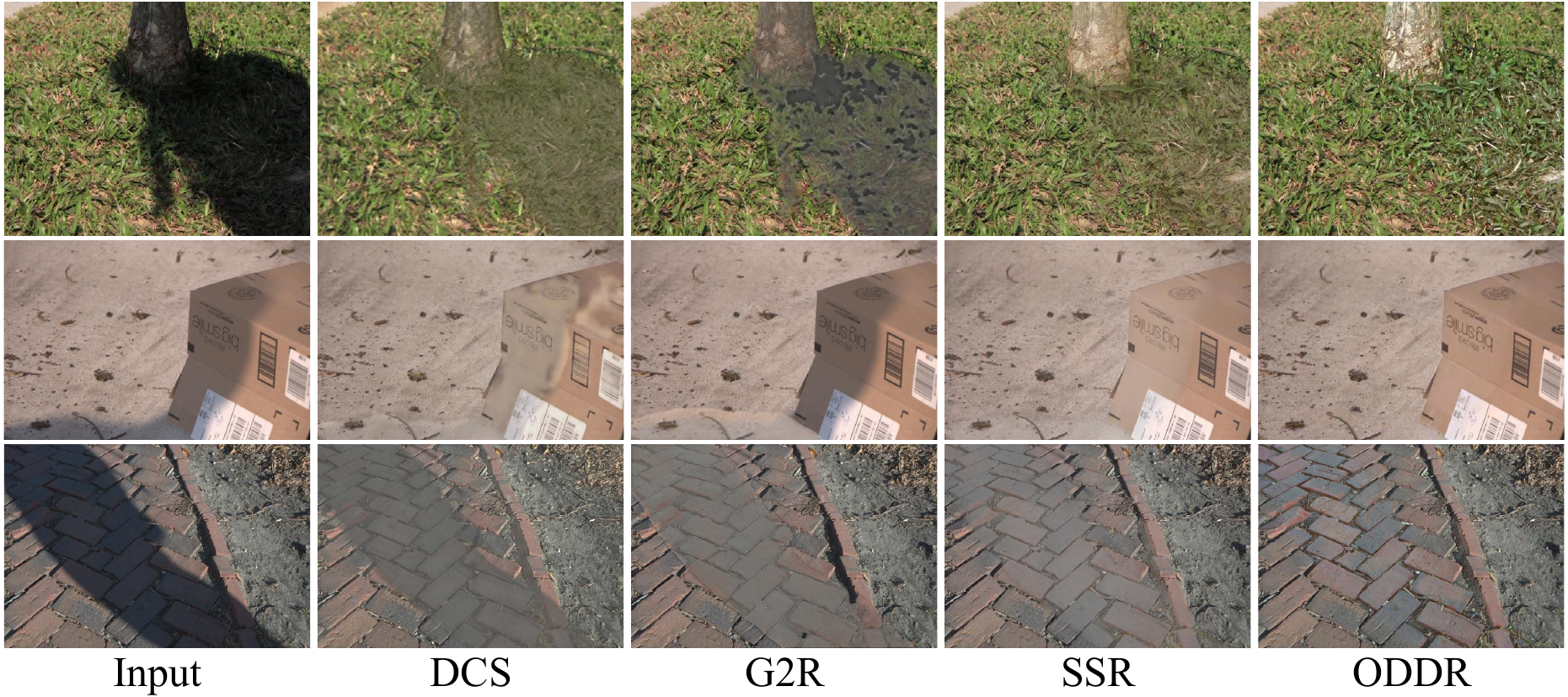}
    \vspace{-1em}
  \caption{Visual comparisons on SRD~\cite{qu2017srd}, LRSS~\cite{gryka2015learninglrss}, and
  UIUC~\cite{guo2012paireduiuc} (top to bottom).}
    \label{fig:unseen_vis_comp}
    \vspace{-1.5em}
\end{figure}

\noindent\textbf{Qualitative Results}
Visual comparisons are presented in Fig.~\ref{fig:comparison} for the AISTD dataset, and in Fig.~\ref{fig:unseen_vis_comp} for the unseen SRD, LRSS, and UIUC datasets. 
On AISTD, methods trained without real-world paired data often reveal dark boundaries or exhibit color suppression around the boundary where a shadow fades into light, whereas our method produces smoother transitions and better tone harmonization across shadow and non-shadow regions.
We observe a similar trend on the unseen SRD, LRSS, and UIUC datasets, where our model surpasses existing approaches in restoring natural illumination and suppressing color artifacts.

\vspace{-0.5em}
\subsection{Ablation Study}
\vspace{-0.5em}

\begin{table}[t]
\centering
\setlength{\tabcolsep}{4pt}
\small
\begin{minipage}[t]{0.43\linewidth}
  \centering
  \caption{Comparison of different reward models and IQA metrics on the AISTD dataset. NR-IQAs denotes the combination of the non-reference IQA models proposed in~\cite{wu2025dp}. ACC denotes the accuracy (\%) of alignment with human preference ranking.}
  \label{tab:reward_comparison}
  \resizebox{\linewidth}{!}{
    \begin{tabular}{l|c|ccc}
    \toprule
    \rowcolor{blue!7}
    Method & \textbf{ACC}$\uparrow$ & \textbf{PSNR}$\uparrow$ & \textbf{SSIM}$\uparrow$ & \textbf{RMSE}$\downarrow$ \\
    \midrule
    Aesthetic Score Predictor & 12.50 & 33.29 & 0.971 & 3.11 \\
    ImageReward & 18.75 & 33.29 & 0.971 & 3.11 \\
    PrefPaint & 25.00 & 33.43 & 0.971 & 3.09 \\
    \midrule
    NR-IQAs & 18.75 & 33.34 & 0.971 & 3.10 \\
    PSNR & 56.25 & -- & -- & -- \\
    \midrule
    \rowcolor{gray!10}
    ShadowReward (ours) & \best{68.75} & \best{34.20} & \best{0.972} & \best{2.97} \\
    \bottomrule
    \end{tabular}
  }
\end{minipage}
\hfill
\begin{minipage}[t]{0.55\linewidth}
  \centering
  \caption{Ablation on cumulative degradations for ShadowReward. Abbreviations: T (Texture distortion), N (Additive noise), C (Chromatic shift), L (Luminance change), B (Boundary artifact).}
  \label{tab:degradation_ablation}
  \resizebox{\linewidth}{!}{
    \begin{tabular}{l|cc|cc}
    \toprule
    \rowcolor{blue!7}
    \multirow{2}{*}{Degradation} &
    \multicolumn{2}{c|}{\textbf{Shadow}} &
    \multicolumn{2}{c}{\textbf{ALL}} \\
    \rowcolor{blue!7}
    \multirow{-2}{*}{Degradation}& PSNR$\uparrow$ & RMSE$\downarrow$ &
    PSNR$\uparrow$ & RMSE$\downarrow$ \\
    \midrule
    T & 36.99 & 6.06 & 33.50 & 3.07 \\
    T, N & 37.17 & 5.95 & 33.61 & 3.05 \\
    T, N, C & 37.82 & 5.70 & 33.96 & 3.01 \\
    T, N, C, L & 38.20 & 5.47 & 34.16 & \best{2.97} \\
    \rowcolor{gray!10}
    T, N, C, L, B & \best{38.37} & \best{5.36} & \best{34.20} & \best{2.97} \\
    \bottomrule
    \end{tabular}
  }
\end{minipage}
\vspace{-1em}
\end{table}

\noindent\textbf{Impact of ShadowReward Model}
To validate the effectiveness of our ShadowReward model, we conduct two complementary evaluations summarized in Tab.~\ref{tab:reward_comparison}.
For comparison, existing reward models such as Aesthetic Score Predictor~\cite{schuhmann2022laion}, ImageReward~\cite{xu2023imagereward}, and PrefPaint~\cite{liu2024prefpaint} are used.
Moreover, IQA metrics are also compared such as NR-IQAs proposed in~\cite{wu2025dp} and PSNR.
We perform an ablation study by substituting ShadowReward in the ODDR training pipeline (Fig.~\ref{fig:main}~(b)) with the aforementioned rewarding methods to compare their deshadowing performance on the AISTD dataset.
Consequently, ODDR fine-tuned with ShadowReward outperforms the alternatives, achieving the best PSNR, SSIM, and RMSE values.
The alignment with human perceptual preferences is further analyzed in the \textbf{Human Preference Alignment} paragraph below.
Note that PSNR-based results are shown as `--' because PSNR relies on ground-truth shadow-free images, which are not used during our training process.

\noindent\textbf{Human Preference Alignment}
To quantitatively assess how accurately ShadowReward reflects human perceptual preferences for shadow removal, we perform an extensive user study involving 23 participants. 
Specifically, these individuals provided preference rankings for images generated by several shadow removal methods~\cite{liu2021shadowLG, xu2025detail, liu2021shadow, kubiak2024s3r}, including our ODD and ODDR models, using the AISTD dataset.
Subsequently, we rank the outcomes based on the predicted scores from each reward model and evaluate how closely these rankings align with human preferences.

In Tab.~\ref{tab:reward_comparison},
the results for accuracy (ACC) confirm that ShadowReward aligns closely with human perceptual preferences.
In contrast, existing reward models and rankings based on a non-reference image quality assessment metric (NR-IQA) exhibit low alignment accuracies (below 30\%), as they rely on generic aesthetic or distortion cues rather than shadow-specific perceptual signals.
Notably, a PSNR-based ranking yields an accuracy of 56.25\% but necessitates ground-truth references for computation. ShadowReward reaches 68.75\% alignment without any ground truth; while the 16-question study is too small to establish a statistically significant advantage over PSNR, it shows that ShadowReward is at least on par with pixel-level comparisons while requiring no reference image (see Appendix Sec.~\ref{supp:human_preference} for inter-annotator agreement).
More details on human evaluation are provided in the Appendix Sec.~\ref{supp:human_preference}.

\noindent\textbf{Effect of Degradations for Human-Annotation-Free Dataset}
Beyond comparing ShadowReward against alternative reward models, we further analyze how the diversity of synthetic corruptions used to train ShadowReward affects its effectiveness.
The types are texture distortion, additive noise, chromatic shift, luminance change, and boundary artifacts, as detailed in Appendix Tab.~\ref{tab:degradation_parameters}.
As shown in Tab.~\ref{tab:degradation_ablation}, incorporating multiple degradations progressively enhances deshadow performance.
Employing all five types of degradation results in the highest PSNR and the lowest RMSE in both Shadow and ALL regions, indicating that a variety of jointly managed corruptions offer more robust supervision for learning.

\noindent\textbf{Conditioning Strategy}
We next examine the architectural choices made in the ODD baseline.
Since our ODD modifies the text-conditioning module of SD, we conduct an ablation study to validate this architectural change in Tab.~\ref{tab:text_conditioning}.
We first input a null prompt to ODD with the text-conditioning module to measure deshadow performance.
We use BLIP~\cite{li2022blip} to extract text from the shadow image, input it to ODD with the text-conditioning module, and measure performance.
We compare these results with our ODD, replacing the text-conditioning with self-attention.
Conditioning on text generated by BLIP results in a slight enhancement compared to using a null prompt, whereas the suggested self-conditioning method (last row) delivers better performance across both Shadow and ALL regions.

\begin{table}[t]
\centering
\setlength{\tabcolsep}{4pt}
\small
\begin{minipage}[t]{0.56\linewidth}
  \centering
  \caption{Results of ODD under different conditioning strategies.}
  \label{tab:text_conditioning}
  \resizebox{\linewidth}{!}{
    \begin{tabular}{l|ccc|ccc}
    \toprule
    \rowcolor{blue!7}
    \multirow{2}{*}{Condition} &
    \multicolumn{3}{c|}{\textbf{Shadow}} &
    \multicolumn{3}{c}{\textbf{ALL}} \\
    \rowcolor{blue!7}
    \multirow{-2}{*}{Condition}
    & \textbf{PSNR$\uparrow$} & \textbf{SSIM$\uparrow$} & \textbf{RMSE$\downarrow$}
    & \textbf{PSNR$\uparrow$} & \textbf{SSIM$\uparrow$} & \textbf{RMSE$\downarrow$} \\
    \midrule
    Null
    & 35.95 & 0.990 & 6.79
    & 32.95 & 0.971 & 3.18 \\
    BLIP
    & 36.29 & 0.990 & 6.54
    & 33.10 & 0.971 & 3.16 \\
    \rowcolor{gray!10}
    ODD (ours)
    & \best{36.67} & \best{0.990} & \best{6.27}
    & \best{33.32} & \best{0.971} & \best{3.11} \\
    \bottomrule
    \end{tabular}
  }
\end{minipage}
\hfill
\begin{minipage}[t]{0.42\linewidth}
  \centering
  \caption{Comparison of efficiency among diffusion shadow removal models.}
  \label{tab:efficiency_comparison}
  \resizebox{\linewidth}{!}{
    \begin{tabular}{l|ccc}
    \toprule
    \rowcolor{blue!7}
    \textbf{Method} & \textbf{NFE$\downarrow$} & \textbf{Inference Time$\downarrow$} & \textbf{FLOPs (G)$\downarrow$} \\
    \midrule
    ShadowDiffusion~\cite{guo2023shadowdiffusion} & 25 & 581.96 & 937.15 \\
    DeS3~\cite{jin2024des3}            & 25 & 538.00 & 812.26 \\
    SSR~\cite{xu2025detail}             & 20 & 680.64 & 2249.50 \\
    \midrule
    \rowcolor{gray!10} ODD (ours)  & \best{1} & \best{113.18} & \best{147.22} \\
    \rowcolor{gray!10} ODDR (ours) & \best{1} & \second{127.41} & \second{158.21} \\
    \bottomrule
    \end{tabular}
  }
\end{minipage}
\vspace{-1.5em}
\end{table}

\noindent\textbf{Computational Cost}
We conduct an ablation on computational efficiency to compare the inference cost with other diffusion-based shadow removal models.
All measurements are performed on a single NVIDIA RTX~4090 GPU with an input resolution of $256\times256$.
As summarized in Tab.~\ref{tab:efficiency_comparison}, multi-step diffusion models~(ShadowDiffusion~\cite{guo2023shadowdiffusion}, DeS3~\cite{jin2024des3}, and SSR~\cite{xu2025detail}) require more than 20 denoising steps, resulting in large FLOPs and slow inference.
Conversely, our single-step models (ODD and ODDR) decrease inference time by over fourfold while preserving similar performance levels.

\vspace{-1em}
\section{Conclusion}
\vspace{-1.0em}

We introduced One-step Deshadow Diffusion via Reward guidance (ODDR), an efficient framework for high-quality shadow removal that requires no real-world paired supervision. Our approach builds upon One-step Deshadow Diffusion (ODD), a baseline that performs single-step shadow-free reconstruction using only synthetic training data, and further refines it through a perceptual reward signal aligned with human preference. At the core of this process lies ShadowReward, a reward model trained on synthetically degraded images that mimic typical failure cases in shadow removal, guiding ODDR toward perceptually pleasing outputs without requiring ground-truth annotations. Extensive experiments demonstrate that our approach achieves state-of-the-art performance among methods without real-world paired supervision, shows strong generalization across diverse datasets, and remains highly efficient compared to multi-step diffusion-based models.
Further analyses on reward-weight sensitivity, shadow-free inputs, and transfer beyond cast shadows are provided in Appendix Secs.~\ref{supp:reward_analysis} and~\ref{supp:transfer}, and a discussion of limitations, representative failure cases, and broader impacts in Appendix Sec.~\ref{supp:limitations}.

\begin{ack}
This research was supported by Culture, Sports and Tourism R\&D Program through the Korea Creative Content Agency grant funded by the Ministry of Culture, Sports and Tourism in 2026 (Project Name: Development of AI-Based Animation Production Technology to Ensure Character and Scene Consistency and Continuity for Enhanced Efficiency and Quality, Project Number: RS-2026-25525207, Contribution Rate: 30\%).
This work was also supported by Institute of Information \& communications Technology Planning \& Evaluation (IITP) grant funded by the Korea government (MSIT) (No.2022-0-00156, Fundamental research on continual meta-learning for quality enhancement of casual videos and their 3D metaverse transformation), and IITP grant funded by the Korea government (MSIT) (No.RS-2020-II201373, Artificial Intelligence Graduate School Program (Hanyang University)).
\end{ack}

\bibliographystyle{plainnat}
\bibliography{main}

\newpage
\input{supple}

\newpage

\end{document}

%% file: supple.tex
\appendix

\section*{Appendix}

This appendix provides additional details and analyses to complement the main paper.
\begin{itemize}
    \item \textbf{Sec.~\ref{supp:human_anno_free}}: Construction of the Human-Annotation-Free Dataset, including degradation parameter settings and the effect of the number of degraded samples.
    \item \textbf{Sec.~\ref{supp:add_experimental_details}}: Detailed hyperparameter configurations used for training ODD, ODDR, and ShadowReward.
    \item \textbf{Sec.~\ref{supp:human_preference}}: Full protocol, statistical reliability, and additional results of the human preference alignment experiment.
    \item \textbf{Sec.~\ref{supp:reward_config}}: Ablation studies on key design choices of ShadowReward, including the feature extractor backbone and spatial reduction strategy.
    \item \textbf{Sec.~\ref{supp:reward_analysis}}: Analysis of reward-guided fine-tuning, including reward attribution and loss-weight sensitivity, reward hacking, robustness to shadow-free inputs, alternative guidance strategies, training data source effects, and non-shadow region color consistency.
    \item \textbf{Sec.~\ref{supp:vae_skip}}: Architecture and ablation of the ODD-Conv design for preserving fine-grained details.
    \item \textbf{Sec.~\ref{supp:transfer}}: Transferability of ShadowReward to ambient lighting normalization and other low-level vision tasks.
    \item \textbf{Sec.~\ref{supp:limitations}}: Limitations of the proposed framework and representative failure cases.
    \item \textbf{Sec.~\ref{supp:licenses}}: Licenses of existing assets used in this work.
    \item \textbf{Sec.~\ref{supp:add_visual_results}}: Additional qualitative results and visual comparisons across multiple datasets.
\end{itemize}

\section{Details of Human-Annotation-Free Dataset}
\label{supp:human_anno_free}

We construct the Human-Annotation-Free Dataset by synthesizing a set of perceptually ordered degraded images for each clean image $y$.
Using a shadow matte $m$ obtained from SynShadow~\cite{inoue2020learning}, we apply degradations to the shadow region of $y$ to simulate unsuccessful shadow removal artifacts.
Specifically, we construct a list of progressively degraded images $\{x^{(j)}\}_{j=0}^{3}$, where $x^{(0)} = y$ represents the clean reference, and each $x^{(j)}$ for $j \ge 1$ applies increasingly severe distortions to the shadow region of $y$.

\begin{table*}[h]
  \centering
  \caption{
  Parameter ranges used to synthesize cumulative degradations (T, N, C, L, B)
  for the Human-Annotation-Free Dataset.
  Higher $j$ corresponds to stronger degradation ($j=0$ is the clean image).
  Here, $r$ denotes the boundary dilation radius,
  $\Delta h$ the hue shift, 
  $s$ the saturation scaling factor, 
  $b$ the brightness scaling factor, 
  and $\gamma$ the luminance gamma parameter.
  All parameters are uniformly sampled within the specified ranges.
  Larger values (or wider ranges) imply more severe distortions.}
  \label{tab:degradation_parameters}
  \resizebox{0.8\linewidth}{!}{%
  \begin{tabular}{c|l|ccc}
    \toprule
    \rowcolor{blue!7}
    \textbf{Type} &
    \textbf{Operation} &
    \textbf{$j=1$} &
    \textbf{$j=2$} &
    \textbf{$j=3$} \\
    \midrule

    T & Gaussian blur &
    $3{\times}3$ &
    $5{\times}5$ &
    $7{\times}7$ \\
    \midrule

    N & Gaussian noise &
    $\sigma{=}0.004$ &
    $\sigma{=}0.008$ &
    $\sigma{=}0.012$ \\
    \midrule

    \multirow{3}{*}{C} & Hue shift &
    $\Delta h{=}\pm0.01$ &
    $\Delta h{=}\pm0.02$ &
    $\Delta h{=}\pm0.03$ \\
    & Saturation scale &
    $s{\in}[0.95,1.05]$ &
    $s{\in}[0.90,1.10]$ &
    $s{\in}[0.85,1.15]$ \\
    & Brightness scale &
    $b{\in}[0.97,1.03]$ &
    $b{\in}[0.94,1.06]$ &
    $b{\in}[0.92,1.08]$ \\
    \midrule

    L & Gamma correction &
    $\gamma{\in}[0.97,1.03]$ &
    $\gamma{\in}[0.94,1.06]$ &
    $\gamma{\in}[0.90,1.10]$ \\
    \midrule

    B & Dilation + Gaussian blur &
    $r{=}3$, $3{\times}3$ &
    $r{=}7$, $5{\times}5$ &
    $r{=}11$, $7{\times}7$ \\

    \bottomrule
  \end{tabular}}%
\end{table*}

\subsection{Degradation Parameter Settings}
To construct a list of degraded images, we define five degradation types that mimic the visual artifacts caused by unsuccessful deshadowing:
texture distortion~(T) applies Gaussian blur to simulate loss of detail,
additive noise~(N) injects Gaussian noise,
chromatic shift~(C) perturbs hue, saturation, and brightness,
luminance change~(L) applies gamma correction,
and boundary artifacts~(B) dilate the shadow boundary followed by Gaussian blur.
All parameters are uniformly sampled within the ranges specified in Tab.~\ref{tab:degradation_parameters}, with level-dependent settings that produce progressively stronger degradations for $j=1,2,3$.

In our experiments, we synthesize each training sample by applying multiple degradation types rather than relying on a single one. 
Depending on the degradation level $j$~(1, 2, or 3), we randomly select 2, 3, or 4 degradation types from (T, N, C, L, B), respectively.
This systematic increase in the number of applied types naturally yields more severe and complex degradations for the higher degradation levels.
This design allows the model to encounter continuously varying and randomly mixed corruption patterns, enabling it to learn more stable and reliable perceptual rankings.

\begin{table}[t]
  \centering
  \caption{
  Impact of the degradation list length $J$ on shadow removal performance.
  }
  \label{tab:shadow_results_abbrev}
  \resizebox{0.6\linewidth}{!}{
  \begin{tabular}{c|cc|cc}
    \toprule
    \rowcolor{blue!7}\multirow{2}{*}{List length ($J$)} & \multicolumn{2}{c|}{\textbf{Shadow}} & \multicolumn{2}{c}{\textbf{All}} \\
    \rowcolor{blue!7}\multirow{-2}{*}{List length ($J$)}& \textbf{PSNR}\,$\uparrow$ & \textbf{RMSE}\,$\downarrow$ & \textbf{PSNR}\,$\uparrow$ & \textbf{RMSE}\,$\downarrow$ \\
    \midrule
    1 & 37.29 & 5.99 & 33.70 & 3.06 \\
    2 & 37.78 & 5.74 & 33.96 & 3.01 \\
    \rowcolor{gray!10}3 &   \best{38.37}     &   \best{5.36}   &    \best{34.20}    &   \best{2.97} \\
    4 &    38.17    &   5.45   &    34.15    &   2.97   \\
    \bottomrule
  \end{tabular}}
\end{table}

\subsection{Length of the Degradation List}
We also analyze how the length of the degradation list~($J$) used to train ShadowReward affects the final shadow removal performance of ODDR.
As shown in Tab.~\ref{tab:shadow_results_abbrev}, the shadow removal performance of ODDR improves as the list length grows from $J=1$ to $J=3$, achieving the best result (38.37\,dB / 5.36 RMSE) at $J=3$.
However, using more than three degraded images leads to a slight drop in performance.
This is primarily because extra distortions push the samples beyond the model's predictive bounds, rendering them irrelevant, while learning too many fine-grained rankings weakens the overall ranking signal.

\section{Additional Experimental Details}
\label{supp:add_experimental_details}

\subsection{Synthetic Shadow Generation}
For ODD training, we generate synthetic shadow images following SynShadow~\cite{inoue2020learning}.
Given a clean image $y$ and a randomly sampled shadow matte $m$, the synthetic shadow image is computed as:
\begin{equation}
  x_{\text{syn}} = (1 - m) \odot y + m \odot (a \cdot y + b),
\end{equation}
where $a < 1$ is a randomly sampled slope controlling shadow darkness, and $b$ is a per-channel intercept.
This affine shadow model produces diverse shadow intensities and color shifts by randomizing $a$ and $b$ for each training pair.
The resulting $(x_{\text{syn}}, y)$ pairs are used to train ODD in a supervised manner.
Additionally, the same shadow matte $m$ is reused during the construction of the Human-Annotation-Free Dataset (Sec.~\ref{supp:human_anno_free}) to apply degradations specifically to the shadow region when training ShadowReward.

\subsection{Loss Hyperparameters}
For ODD training (Eq.~(\ref{eq:loss_odd})), we use the LPIPS perceptual loss~\cite{zhang2018unreasonable} with a weighting factor of $\lambda_{\text{per}} = 2$.
When training ODDR (Eq.~(\ref{eq:loss_oddr})), we additionally apply the alignment loss with $\lambda_{\text{align}} = 2$, and set the reward-guidance weight to $\lambda_{r} = 1\times10^{-1}$.
For ShadowReward, we train the ListMLE-based ranking model~\cite{xia2008listwise} (Eq.~(\ref{eq:loss_list})) using a temperature parameter of $\tau = 1$.
All hyperparameters are empirically chosen and kept fixed across all experiments.

\subsection{LoRA Configuration}
We use LoRA~\cite{hu2022lora} with rank 16 and Gaussian initialization to adapt the frozen Stable Diffusion UNet.
LoRA modules are inserted into both the attention layers (\texttt{to\_k}, \texttt{to\_q}, \texttt{to\_v}, \texttt{to\_out}) and the convolutional layers (\texttt{conv1}, \texttt{conv2}, \texttt{conv\_in}, \texttt{conv\_shortcut}, \texttt{conv\_out}), as well as the feed-forward layers (\texttt{ff.net}) and projection layers (\texttt{proj\_in}, \texttt{proj\_out}).
For ODD training, both ODD-LoRA and ODD-Conv are optimized while the original SD weights remain frozen.
For ODDR fine-tuning, a separate ODDR-LoRA module is initialized and optimized while ODD-LoRA and ODD-Conv are frozen.

\subsection{ShadowReward Architecture Details}
We use DINOv2 ViT-B/14~\cite{oquab2023dinov2} (patch size 14, embedding dimension 768) as the feature extractor, which is kept frozen during training.
The convolutional head $\psi_\theta$ takes as input the concatenation of the shadow tokens, the corrupted image tokens, their element-wise difference, and its absolute value, resulting in an input dimension of $768 \times 4 = 3072$ channels.
The head consists of: a $1\times1$ projection layer ($3072 \rightarrow 256$), a $3\times3$ depthwise separable convolution ($256 \rightarrow 128$, groups$=$128), a $1\times1$ pointwise convolution ($128 \rightarrow 64$), Dropout2d, and a final $1\times1$ convolution ($64 \rightarrow 1$) followed by a sigmoid activation.
GELU activation is applied after each intermediate layer.

\subsection{Data Augmentation}
During training, we apply the following augmentations to the input images: (1) reflection padding to ensure a minimum resolution of $512\times512$; (2) random cropping to $512\times512$; (3) random horizontal flipping with probability 0.5; and (4) random rotation sampled from $\{90^{\circ}, 180^{\circ}, 270^{\circ}\}$.
These augmentations are applied consistently across all training stages (ODD, ODDR, and ShadowReward).

\subsection{Inference Details}
At inference time, the input image is first padded using reflection padding so that both spatial dimensions are multiples of 8, which is required by the VAE encoder.
The model then performs a single-step denoising pass at timestep $t{=}999$ without classifier-free guidance (CFG)~\cite{ho2022classifier}.
The output is cropped back to the original input resolution after decoding.

\subsection{Training Cost}
All experiments are conducted on a single NVIDIA RTX A6000 GPU.
ODD training takes approximately 24 hours, while ODDR reward-guided fine-tuning converges in approximately 12 hours.

\section{Observations on Human Preference Alignment}
\label{supp:human_preference}

\begin{figure}
    \centering
    \includegraphics[width=0.9\linewidth]{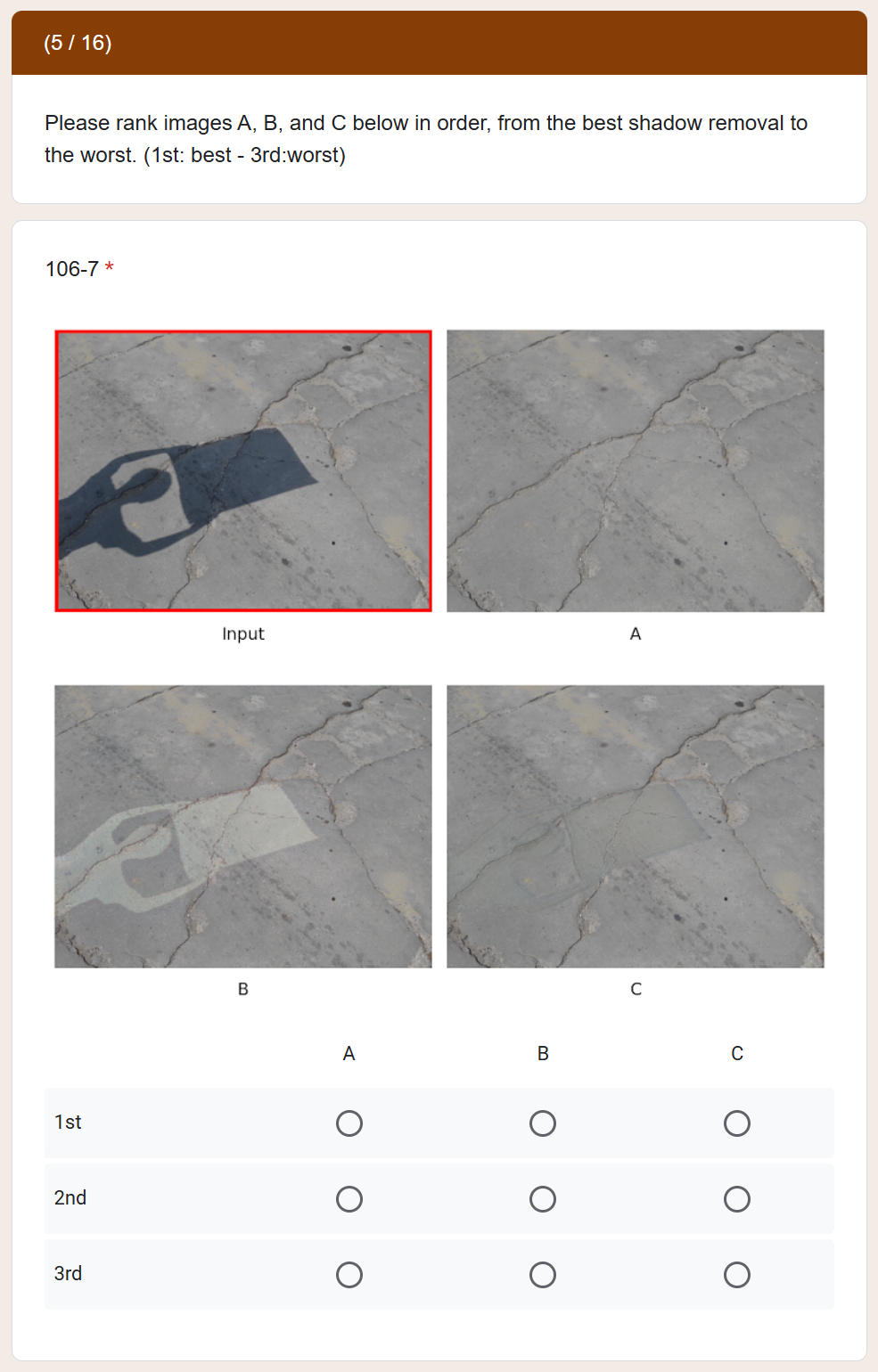}
    \caption{User study interface for human preference acquisition. Annotators rank three randomized results (A, B, C) against the input image (red border).}
    \label{fig:survey_example}
\end{figure}

\subsection{User Study Details}
To measure how well ShadowReward aligns with human preference, we conducted a user study collecting human rankings for shadow removal results.
The survey interface, shown in Fig.~\ref{fig:survey_example}, was designed to facilitate direct visual comparison.
\begin{itemize}
    \item \textbf{Survey Structure / Interface}:
    As illustrated in Fig.~\ref{fig:survey_example}, each question presented a $2 \times 2$ grid layout containing one input shadow image and three differently deshadowed results (labeled A, B, and C), where the three results were randomly selected from different shadow removal methods.
    \item \textbf{Ranking Task}:
    Participants were instructed to ``rank images A, B, and C from the most effective shadow removal to the least effective.''
    We utilized a matrix-style selection interface, ensuring that annotators explicitly assigned a unique rank (1st, 2nd, or 3rd) to each candidate without duplication.
    \item \textbf{Data Source}:
    The 16 input images were randomly selected from the AISTD test dataset~\cite{le2019shadow}.
    The three comparison images for each question were sampled from a pool of six different models: four existing deshadowing models~\cite{liu2021shadowLG, xu2025detail, liu2021shadow, kubiak2024s3r} and our proposed ODD and ODDR models.
    The positions of the candidate images (A, B, C) were randomized for each question to prevent positional bias.
    \item \textbf{Participants}:
    A total of 23 annotators participated in the survey.
    The participant pool consisted of a diverse group, ranging from low-level vision experts (graduate students and Ph.D. researchers) to general users without technical background.
    This diversity ensures that the evaluation reflects both technical precision and general perceptual quality.
    \item \textbf{Ground Truth Establishment}:
    The ground-truth ranking for each question was determined by aggregating the responses via majority voting.
    This consensus ranking was subsequently used to calculate the reward model accuracy reported in Tab.~\ref{tab:reward_comparison}.
    \item \textbf{Evaluation Setting}:
    Since every input is a real AISTD shadow image and every candidate is the output of a real shadow removal model, the reported accuracy measures ShadowReward on (real shadow, real model output) pairs, i.e., the setting in which it is used during ODDR fine-tuning, rather than on the synthetic degradations it is trained on.
    \item \textbf{Statistical Reliability}:
    To verify that the human consensus is a reliable reference, we measure the agreement of individual responses with the majority-vote ranking: annotators agree on 79.6\% of full three-way rankings and on 86.7\% of top choices, well above chance (1/6 for a full ranking).
    We note that the study contains only 16 questions; ShadowReward's accuracy (68.75\%, 11/16) is therefore not statistically separable from PSNR-based ranking (56.25\%, 9/16), and we interpret it as being on par with PSNR while requiring no ground truth.
    In contrast, the margin over existing reward models and NR-IQA metrics (all below 30\%) is substantial.
\end{itemize}

\begin{figure*}
    \centering
    \includegraphics[width=0.95\linewidth]{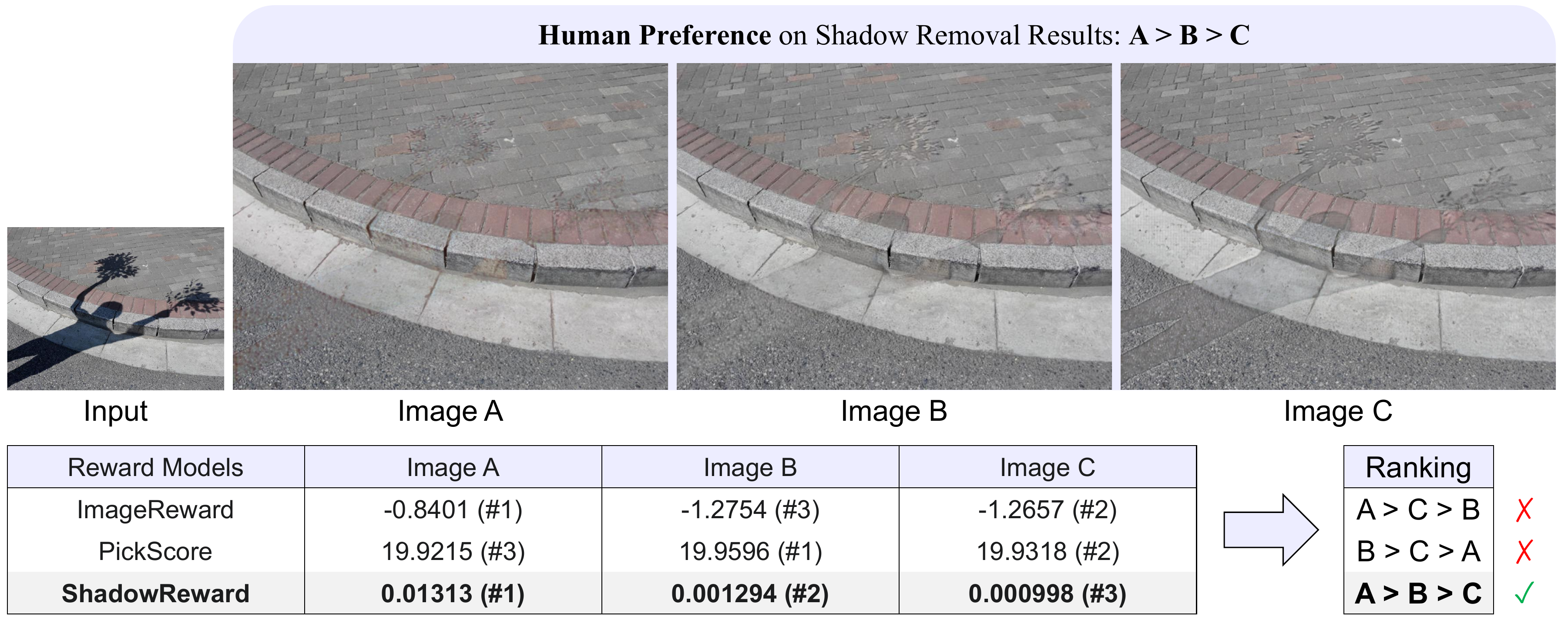}
    \caption{\textbf{Comparison of reward model ranking accuracy.}
     ShadowReward correctly ranks the shadow removal results in alignment with human preference ($A>B>C$), while existing reward models fail to match human perception.
    }
    \label{fig:human_pref}
\end{figure*}

\subsection{Ranking Accuracy Comparison}
To demonstrate that ShadowReward better captures human preference for shadow removal quality, we compare it against existing reward models commonly used in image generation:
ImageReward~\cite{xu2023imagereward} and PickScore~\cite{kirstain2023pick}.
As shown in Fig.~\ref{fig:human_pref}, general-purpose reward models do not align with human rankings for shadow removal results.
They often favor images with stronger global semantics, disregarding noticeable residual shadows or color distortions. 
In contrast, ShadowReward consistently aligns with human-preferred ordering, accurately prioritizing restorations that feature clean shadow removal and coherent illumination. 
This demonstrates that ShadowReward is specifically well-aligned with the requirements of shadow removal, providing task-aware perceptual judgments that general-purpose VLM-based reward models cannot achieve.

\begin{figure}
    \centering
    \includegraphics[width=\linewidth]
    {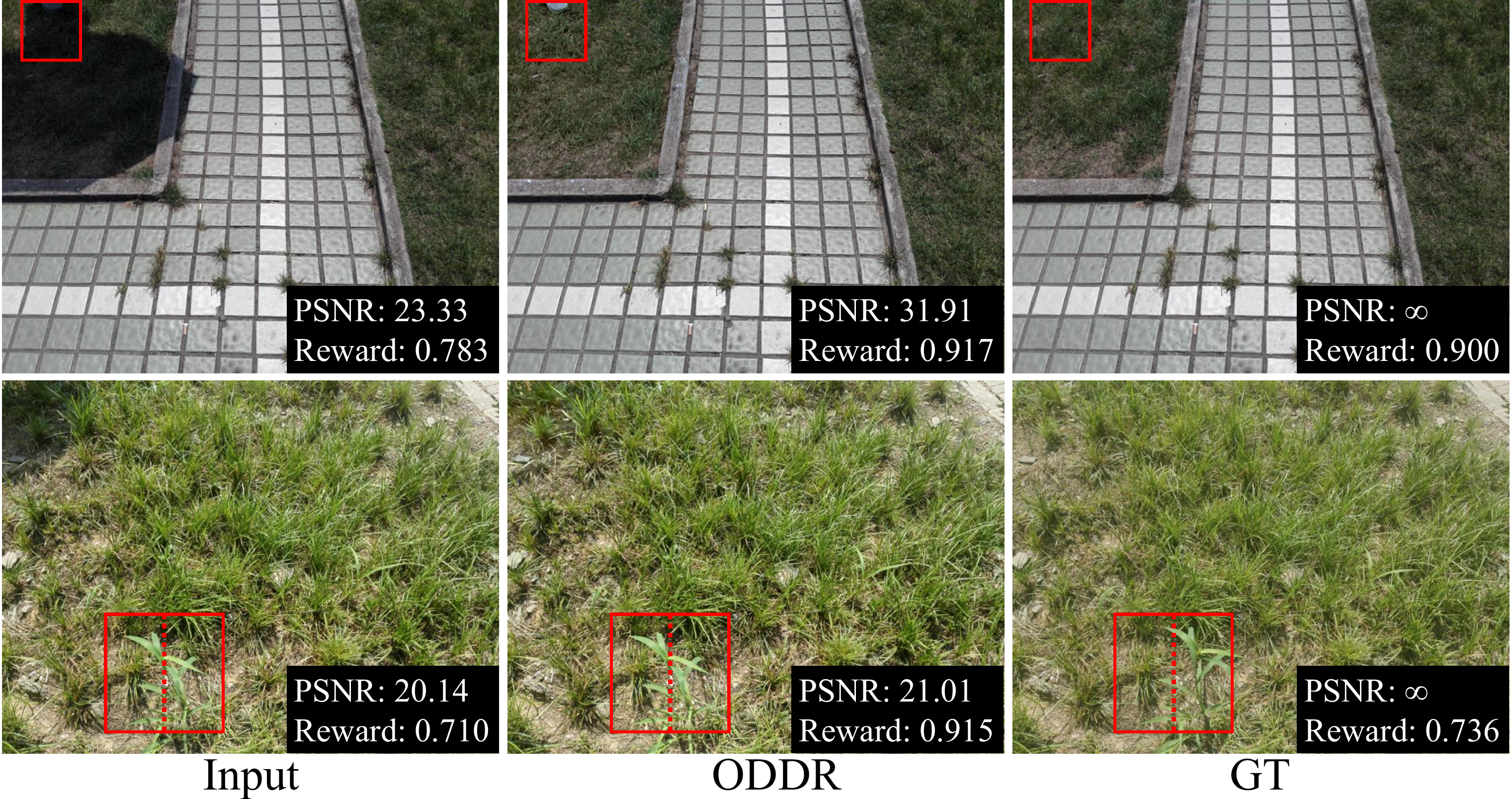}
    \caption{
    PSNR is unreliable under GT misalignment, while our ShadowReward (labeled `Reward') provides perceptually consistent evaluation.
}
    \label{fig:RewardModel}
\end{figure}

\subsection{Perceptual Evaluation Beyond PSNR}
Traditional metrics such as PSNR depend on strict pixel-level correspondence with paired ground truth and therefore become unreliable when misalignment occurs between two images.
As illustrated in Fig.~\ref{fig:RewardModel}, mismatches stemming from transient objects (e.g., the shoe in the top row) or environmental movements (e.g., shifting grass in the bottom row) lead to unfairly low PSNR scores for high-quality restorations, even when the shadows are correctly removed.
In contrast, ShadowReward provides a more perceptually reliable assessment by focusing on visual coherence rather than pixel-wise similarity. 
It assigns higher scores to restorations with natural tone and structure, and penalizes outputs with residual shadows or color inconsistency, regardless of ground-truth bias.
These examples highlight that PSNR can fail to reflect human perceptual preference under misaligned conditions, whereas ShadowReward, which requires no ground truth, remains unaffected by such misalignment and thus serves as a useful complementary perceptual evaluator for shadow removal.

\section{Ablation Study of ShadowReward}
\label{supp:reward_config}
We conduct ablation studies on key design choices of ShadowReward, including the feature extractor backbone and the spatial reduction strategy.

\begin{table}[t]
  \centering
  \caption{
  Impact of different feature extractor backbones for ShadowReward.
  }
  \label{tab:shadow_results_feature}
  \resizebox{0.55\linewidth}{!}{
  \begin{tabular}{c|cc|cc}
    \toprule
    \rowcolor{blue!7}\multirow{2}{*}{\textbf{Backbone}} & \multicolumn{2}{c|}{\textbf{Shadow}} & \multicolumn{2}{c}{\textbf{All}} \\
    \rowcolor{blue!7}\multirow{-2}{*}{\textbf{Backbone}}& \textbf{PSNR}\,$\uparrow$ & \textbf{RMSE}\,$\downarrow$ & \textbf{PSNR}\,$\uparrow$ & \textbf{RMSE}\,$\downarrow$ \\
    \midrule
    CLIP & 37.27 & 5.91 & 33.67 & 3.04 \\
    MoCov3 & 37.72 & 5.71 & 33.91 & 3.01 \\
    \rowcolor{gray!10}DINOv2  &   \best{38.37}     &   \best{5.36}   &    \best{34.20}    &   \best{2.97}   \\
    \bottomrule
  \end{tabular}}
\end{table}

\subsection{Feature Extractor}
We investigate the impact of different feature extractors used in ShadowReward. 
As shown in Tab.~\ref{tab:shadow_results_feature}, the reward model with DINOv2~\cite{oquab2023dinov2} performs better than those with CLIP~\cite{radford2021learning} and MoCov3~\cite{chen2021empirical}, achieving a shadow-region PSNR of 38.37 and an RMSE of 5.36, compared to 37.27/5.91 for CLIP and 37.72/5.71 for MoCov3.
This indicates that DINOv2’s superior capability in capturing local structural dependencies and illumination variations makes it more effective for evaluating shadow boundaries and color consistency.

\begin{table}[t]
  \centering
  \caption{
  Impact of different spatial reduction methods for ShadowReward.
  }
  \label{tab:caption_conditions}
  \resizebox{0.6\linewidth}{!}{
  \begin{tabular}{c|cc|cc}
    \toprule
    \rowcolor{blue!7}\multirow{2}{*}{\textbf{Method}} & \multicolumn{2}{c|}{\textbf{Shadow}} & \multicolumn{2}{c}{\textbf{All}} \\
    \rowcolor{blue!7}\multirow{-2}{*}{\textbf{Method}}& \textbf{PSNR}\,$\uparrow$ & \textbf{RMSE}\,$\downarrow$ & \textbf{PSNR}\,$\uparrow$ & \textbf{RMSE}\,$\downarrow$ \\
    \midrule
    Mean & 37.66 & 5.77 & 33.90 & 3.02 \\
    Max & 37.62 & 5.56 & 33.89 & 3.02 \\
    Top-$K$(0.05)  &   38.28     &   5.48   &    34.19    &   \best{2.97}   \\
    Top-$K$(0.15) &    38.08    &   5.51   &    34.12    &   2.98   \\
    \midrule 
    \rowcolor{gray!10}Top-$K$(0.1) &   \best{38.37}     &   \best{5.36}   &    \best{34.20}    &   \best{2.97}   \\
    \bottomrule
  \end{tabular}}
\end{table}

\subsection{Spatial Reduction Method}
We further evaluate different spatial reduction strategies used for aggregating patch-level scores into the final perceptual reward.
Tab.~\ref{tab:caption_conditions} shows that simple mean or max pooling methods underperform compared to Top-$K$ pooling.
Among the variants, Top-$K$ pooling with $K{=}0.1$ (top 10\% of spatial responses) yields the best balance between localization and robustness, achieving a Shadow PSNR of 38.37\,dB while maintaining an overall RMSE of 2.97.
This demonstrates that selectively averaging the most informative spatial responses allows the reward model to focus on shadow boundaries while suppressing noise from irrelevant regions.

\subsection{ListMLE Temperature}
We tested $\tau \in \{0.5, 1.0, 1.5\}$ and observed pairwise preference accuracies of $0.939$, $0.969$, and $0.963$ respectively on the AISTD test set (clean GT vs.\ ODD output), indicating that ShadowReward is robust to moderate variations in the ListMLE temperature; we therefore use $\tau{=}1.0$ following standard practice~\cite{xia2008listwise}.

\section{Analysis of Reward-Guided Fine-Tuning}
\label{supp:reward_analysis}

In this section, we provide further analyses to validate the design choices and effectiveness of our reward-guided fine-tuning strategy.

\begin{table}[t]
\centering
\caption{Reward attribution and reward-weight sensitivity on AISTD. $\lambda_r{=}0$ corresponds to anchor-only fine-tuning. $\lambda_{\text{align}}{=}2$ in all rows.}
\label{tab:reward_weight}
\resizebox{\linewidth}{!}{
\begin{tabular}{l|ccc|ccc|ccc}
\toprule
\rowcolor{blue!7}
 & \multicolumn{3}{c|}{\textbf{Shadow}} & \multicolumn{3}{c|}{\textbf{Non-Shadow}} & \multicolumn{3}{c}{\textbf{ALL}} \\
\rowcolor{blue!7}
$\lambda_r$ & PSNR$\uparrow$ & SSIM$\uparrow$ & RMSE$\downarrow$ & PSNR$\uparrow$ & SSIM$\uparrow$ & RMSE$\downarrow$ & PSNR$\uparrow$ & SSIM$\uparrow$ & RMSE$\downarrow$ \\
\midrule
ODD (base)        & 36.67 & 0.990 & 6.27 & 37.45 & 0.984 & 2.49 & 33.32 & 0.971 & 3.11 \\
0 (anchor only)   & 36.63 & 0.990 & 6.29 & 37.45 & 0.984 & 2.50 & 33.30 & 0.971 & 3.11 \\
0.01              & 36.91 & 0.991 & 6.13 & 37.45 & 0.984 & 2.48 & 33.47 & 0.971 & 3.08 \\
\rowcolor{gray!10}
0.1 (ours)        & 38.37 & 0.991 & 5.36 & 37.34 & 0.984 & 2.50 & 34.20 & 0.972 & 2.97 \\
0.5               & 38.34 & 0.991 & 5.41 & 37.30 & 0.984 & 2.51 & 34.16 & 0.972 & 2.99 \\
1.0               & 38.32 & 0.991 & 5.44 & 37.27 & 0.984 & 2.53 & 34.12 & 0.971 & 3.00 \\
10.0              & 38.29 & 0.991 & 5.52 & 37.26 & 0.984 & 2.54 & 34.01 & 0.971 & 3.02 \\
100.0             & 38.48 & 0.991 & 5.36 & 36.77 & 0.983 & 2.63 & 33.87 & 0.970 & 3.07 \\
\bottomrule
\end{tabular}}
\end{table}

\subsection{Reward Attribution and Reward-Weight Sensitivity}
\label{supp:reward_weight}
A natural question is whether the ODD$\rightarrow$ODDR improvement originates from the reward or from self-distillation toward the frozen ODD pseudo-target.
Both alignment terms in Eq.~(\ref{eq:loss_oddr}), the $\ell_2$ distance and the perceptual distance to $\hat{y}_{\text{ODD}}$, attain their minimum of zero exactly at the ODD output, i.e., at initialization.
Consequently, with $\lambda_r{=}0$ the objective is already minimized and fine-tuning leaves the model essentially unchanged, while for $\lambda_r{>}0$ the reward is the only term that moves the model beyond ODD.
Tab.~\ref{tab:reward_weight} confirms this empirically: the anchor-only setting ($\lambda_r{=}0$) reproduces ODD (33.30 vs.\ 33.32\,dB ALL PSNR), so the entire gain of ODDR (+1.70\,dB shadow-region PSNR) is attributable to the reward.
Note that the small value of $\lambda_r$ relative to $\lambda_{\text{align}}$ does not imply a small contribution, since the two weights multiply loss terms that live on different scales.

Performance is stable over a wide range of reward weights: ALL PSNR remains between 34.01 and 34.20\,dB for $\lambda_r \in [0.1, 10]$, with $\lambda_r{=}0.1$ being the best setting.

\noindent\textbf{Reward hacking.}
To locate where over-optimization begins, we further increase the weight to $\lambda_r{=}100$.
Here, the shadow-region PSNR reaches its highest value (38.48\,dB), but the non-shadow and ALL PSNR drop markedly (37.34$\rightarrow$36.77\,dB and 34.20$\rightarrow$33.87\,dB): the reward keeps improving the region it targets while overall reconstruction quality degrades, the typical signature of reward hacking.
Our default $\lambda_r{=}0.1$ lies well within the stable regime, where the alignment terms protect non-shadow fidelity.

\begin{table}[t]
\centering
\caption{Alignment-weight sensitivity on AISTD ($\lambda_r{=}0.1$ in all rows).}
\label{tab:align_weight}
\resizebox{\linewidth}{!}{
\begin{tabular}{l|ccc|ccc|ccc}
\toprule
\rowcolor{blue!7}
 & \multicolumn{3}{c|}{\textbf{Shadow}} & \multicolumn{3}{c|}{\textbf{Non-Shadow}} & \multicolumn{3}{c}{\textbf{ALL}} \\
\rowcolor{blue!7}
$\lambda_{\text{align}}$ & PSNR$\uparrow$ & SSIM$\uparrow$ & RMSE$\downarrow$ & PSNR$\uparrow$ & SSIM$\uparrow$ & RMSE$\downarrow$ & PSNR$\uparrow$ & SSIM$\uparrow$ & RMSE$\downarrow$ \\
\midrule
0            & 38.17 & 0.991 & 5.54 & 37.32 & 0.984 & 2.52 & 34.07 & 0.971 & 3.01 \\
1.0          & 38.27 & 0.991 & 5.48 & 37.27 & 0.984 & 2.53 & 34.09 & 0.971 & 3.01 \\
\rowcolor{gray!10}
2.0 (ours)   & 38.37 & 0.991 & 5.36 & 37.34 & 0.984 & 2.50 & 34.20 & 0.972 & 2.97 \\
5.0          & 37.36 & 0.991 & 5.87 & 37.45 & 0.984 & 2.48 & 33.74 & 0.972 & 3.04 \\
\bottomrule
\end{tabular}}
\end{table}

\subsection{Alignment-Weight Sensitivity}
\label{supp:align_weight}
Tab.~\ref{tab:align_weight} varies the alignment weight $\lambda_{\text{align}}$.
ALL PSNR remains stable for $\lambda_{\text{align}} \in \{0, 1, 2\}$ (34.07 / 34.09 / 34.20\,dB).
At $\lambda_{\text{align}}{=}5$, the anchor dominates the reward and the shadow-region PSNR drops (38.37$\rightarrow$37.36\,dB), the expected trade-off of over-weighting the anchor.
We therefore use $\lambda_{\text{align}}{=}2$.

\begin{table}[t]
\centering
\caption{Robustness to shadow-free inputs. The 540 shadow-free images of the AISTD test set are fed to each model, and the output is compared with the clean input. Higher PSNR and lower RMSE indicate better preservation.}
\label{tab:shadow_free}
\resizebox{0.5\linewidth}{!}{
\begin{tabular}{l|cc}
\toprule
\rowcolor{blue!7}
Shadow-free input & PSNR$\uparrow$ & RMSE$\downarrow$ \\
\midrule
SSR~\cite{xu2025detail} & 40.08 & 2.08 \\
\rowcolor{gray!10}
ODDR (ours) & \best{45.10} & \best{0.83} \\
\bottomrule
\end{tabular}}
\end{table}

\subsection{Robustness to Shadow-Free and Soft Shadows}
\label{supp:shadow_free}
A shadow removal model should not alter content when there is little or nothing to remove.
To test this, we pass the 540 shadow-free images of the AISTD test set through the model and compare each output with its clean input.
As shown in Tab.~\ref{tab:shadow_free}, ODDR leaves clean inputs nearly unchanged (45.10\,dB PSNR), well above the strongest baseline SSR (40.08\,dB), indicating that it does not erase genuine detail in the absence of shadows.
This is consistent with the non-shadow preservation observed in Tab.~\ref{tab:shadow_results} and Fig.~\ref{fig:comparison}.
For faint shadows, LRSS~\cite{gryka2015learninglrss} consists of soft shadows, and ODDR achieves the best results on it (Tab.~\ref{tab:lrss_uiuc_results}), showing that it handles soft shadow boundaries without over-editing.

\begin{table}[t]
\centering
\caption{Comparison with alternative guidance strategies on the AISTD dataset. All methods use the same ODD baseline and unpaired fine-tuning protocol.}
\label{tab:alternative_guidance}
\resizebox{0.70\linewidth}{!}{
\begin{tabular}{l|cc|cc}
\toprule
\rowcolor{blue!7}
\multirow{2}{*}{\textbf{Guidance Method}} &
\multicolumn{2}{c|}{\textbf{Shadow}} &
\multicolumn{2}{c}{\textbf{ALL}} \\
\rowcolor{blue!7}
\multirow{-2}{*}{\textbf{Guidance Method}} &
\textbf{PSNR}$\uparrow$ & \textbf{RMSE}$\downarrow$ &
\textbf{PSNR}$\uparrow$ & \textbf{RMSE}$\downarrow$ \\
\midrule
Shadow Detection & 36.69 & 6.26 & 33.33 & 3.11 \\
Conditional GAN & 36.79 & 6.40 & 33.36 & 3.12 \\
DINO Discriminator & 37.79 & 5.74 & 33.91 & 3.03 \\
\midrule
\rowcolor{gray!10} ShadowReward (ours) & \best{38.37} & \best{5.36} & \best{34.20} & \best{2.97} \\
\bottomrule
\end{tabular}}
\end{table}

\subsection{Comparison with Alternative Guidance Strategies}
To justify the use of ShadowReward over other possible guidance signals, we compare it against three alternative methods under the same unpaired fine-tuning protocol used for ODDR: (1)~a SAM-based shadow detection model~\cite{chen2023sam}, which penalizes the shadow detection confidence in the restored output to encourage shadow-free results; (2)~a conditional GAN discriminator~\cite{isola2017image}, trained to distinguish unpaired shadow-free images from the restored images; and (3)~a DINO-based discriminator adapted for one-step diffusion in low-level vision tasks.

As shown in Tab.~\ref{tab:alternative_guidance}, ShadowReward consistently outperforms all alternatives.
The shadow detection approach only marginally improves over ODD, as it provides a binary shadow presence signal rather than fine-grained perceptual feedback.
The conditional GAN discriminator suffers from unstable adversarial training dynamics in the unpaired setting and occasionally introduces new artifacts.
The DINO-based discriminator performs reasonably well due to its strong structural and perceptual cue-capturing capabilities, yet it still falls short of ShadowReward by 0.58\,dB in shadow-region PSNR.
These results demonstrate that learning a preference ranking over restored results, as done by ShadowReward, provides a more effective supervision signal than discriminative or detection-based alternatives.

\begin{table}[t]
\centering
\caption{Effect of training data source overlap on the AISTD dataset. USR denotes a disjoint shadow-free dataset used to train ODD.}
\label{tab:data_overlap}
\resizebox{0.70\linewidth}{!}{
\begin{tabular}{l|c|cc|cc}
\toprule
\rowcolor{blue!7}
\multirow{2}{*}{\textbf{Method}} &
\multirow{2}{*}{\textbf{ODD Train}} &
\multicolumn{2}{c|}{\textbf{Shadow}} &
\multicolumn{2}{c}{\textbf{ALL}} \\
\rowcolor{blue!7}
\multirow{-2}{*}{\textbf{Method}} &
\multirow{-2}{*}{\textbf{ODD Train}} &
\textbf{PSNR}$\uparrow$ & \textbf{RMSE}$\downarrow$ &
\textbf{PSNR}$\uparrow$ & \textbf{RMSE}$\downarrow$ \\
\midrule
ODD  & AISTD & 36.67 & 6.27 & 33.32 & 3.11 \\
\rowcolor{gray!10} ODDR & AISTD & \best{38.37} & \best{5.36} & \best{34.20} & \best{2.97} \\
\midrule
ODD  & USR   & 36.08 & 6.70 & 32.86 & 3.24 \\
\rowcolor{gray!10} ODDR & USR   & 37.58 & 5.80 & 33.71 & 3.13 \\
\bottomrule
\end{tabular}}
\end{table}

\subsection{Effect of Training Data Source}
\label{supp:data_overlap}
A potential concern regarding data overlap is that ODD and ODDR both rely on the AISTD dataset: ODD uses its shadow-free images to synthesize training pairs, while ODDR uses its real shadow images for reward-guided fine-tuning. 
To verify that the improvement of ODDR is not simply due to this overlapping scene distribution, we train ODD using shadow-free images from a completely disjoint dataset, USR~\cite{hu2019mask}, while keeping ODDR fine-tuning on AISTD real shadow images as in the standard pipeline.

As shown in Tab.~\ref{tab:data_overlap}, even when ODD is trained on a different data source, ODDR still yields consistent improvements by raising the ALL PSNR from 32.86\,dB to 33.71\,dB, confirming that the gains stem from reward-guided optimization rather than reusing the same scene distribution.
This result validates the generality of our reward-guided fine-tuning approach and demonstrates that ShadowReward provides an effective learning signal regardless of the baseline training data.

\subsection{Why Does ODDR Preserve Non-Shadow Region Colors?}

\begin{figure}[t!]
    \centering
    \includegraphics[width=0.85\linewidth]{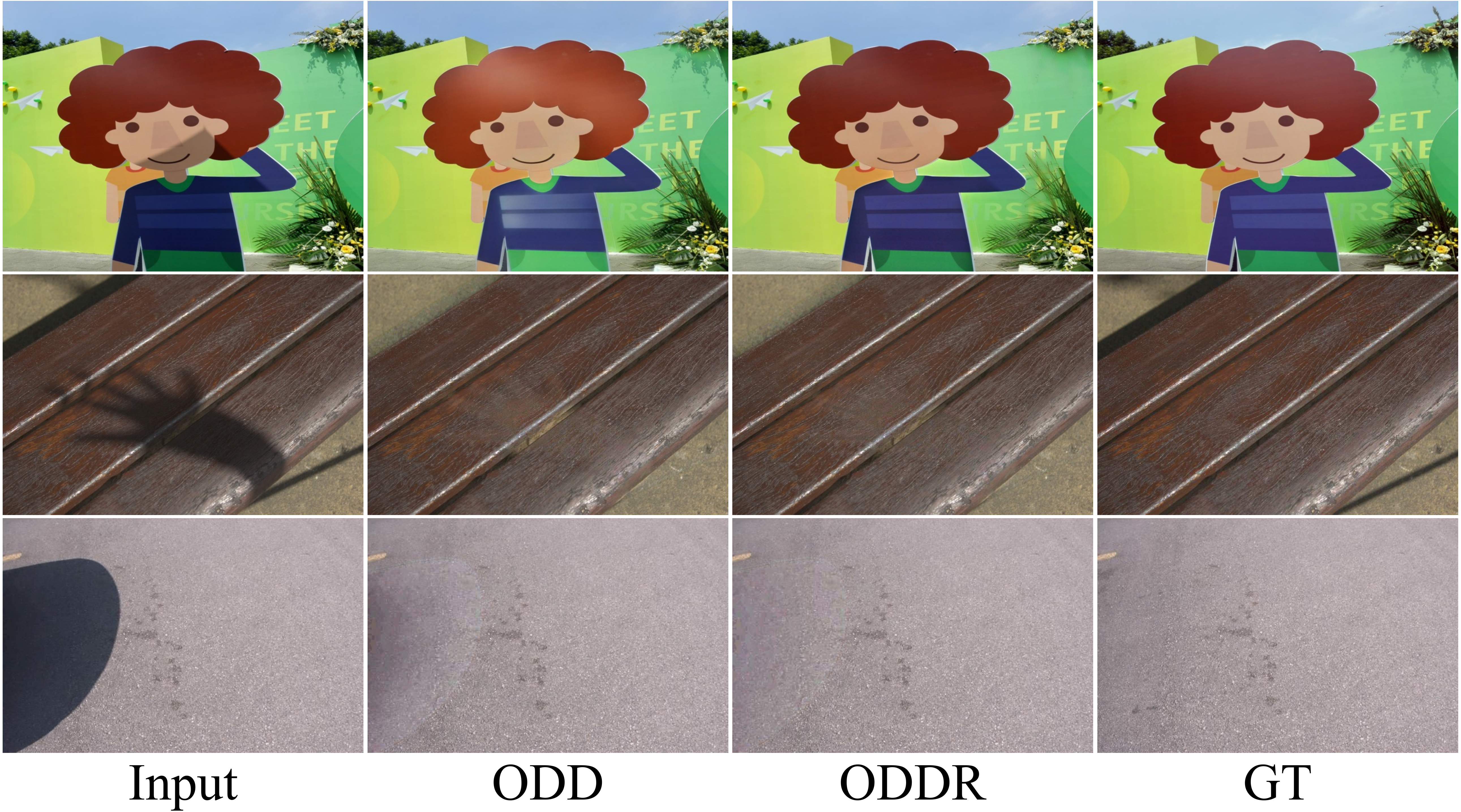}
    \vspace{-0.5em}
    \caption{Visual comparisons of our ODD and ODDR on the AISTD~\cite{le2019shadow} (top), LRSS~\cite{gryka2015learninglrss} (middle), and UIUC~\cite{guo2012paireduiuc} (bottom) datasets.}
    \label{fig:odd_oddr_vis_comp}
\end{figure}

To visually evaluate the effectiveness of the ShadowReward model across different datasets, we provide visual comparisons of ODD and ODDR in Fig.~\ref{fig:odd_oddr_vis_comp}.
As shown in the first row, ODDR removes shadows more completely within the shadow regions, producing cleaner illumination and more natural color transitions compared to ODD.
Moreover, ODDR restores boundary areas more cleanly in the second and last rows, showing fewer artifacts and better consistency between shadow and non-shadow regions.
These findings show that our learned reward signal effectively guides the model to produce perceptually improved shadow-free reconstructions.

Beyond shadow-region quality, we also observe that ODD occasionally introduces undesirable color shifts in non-shadow regions, whereas ODDR effectively preserves the original color tone.
This improvement can be attributed to the design of ShadowReward, which evaluates the overall perceptual quality of the restored image by comparing it against the shadow input.
Since ShadowReward is trained to penalize perceptually unnatural artifacts, including chromatic distortions and luminance inconsistencies, it provides a gradient signal that discourages global color drift during ODDR fine-tuning.
Combined with the alignment loss ($\mathcal{L}_{\text{per}}$ and $\ell_2$) that anchors ODDR outputs to the ODD pseudo-target, the reward term selectively improves shadow regions while suppressing unintended modifications in non-shadow areas.
This complementary supervision explains why ODDR achieves better color consistency across the entire image.

\section{ODD-Conv: Skip-Connection Design in the VAE}
\label{supp:vae_skip}

\begin{figure}
    \centering
    \includegraphics[width=0.7\linewidth]{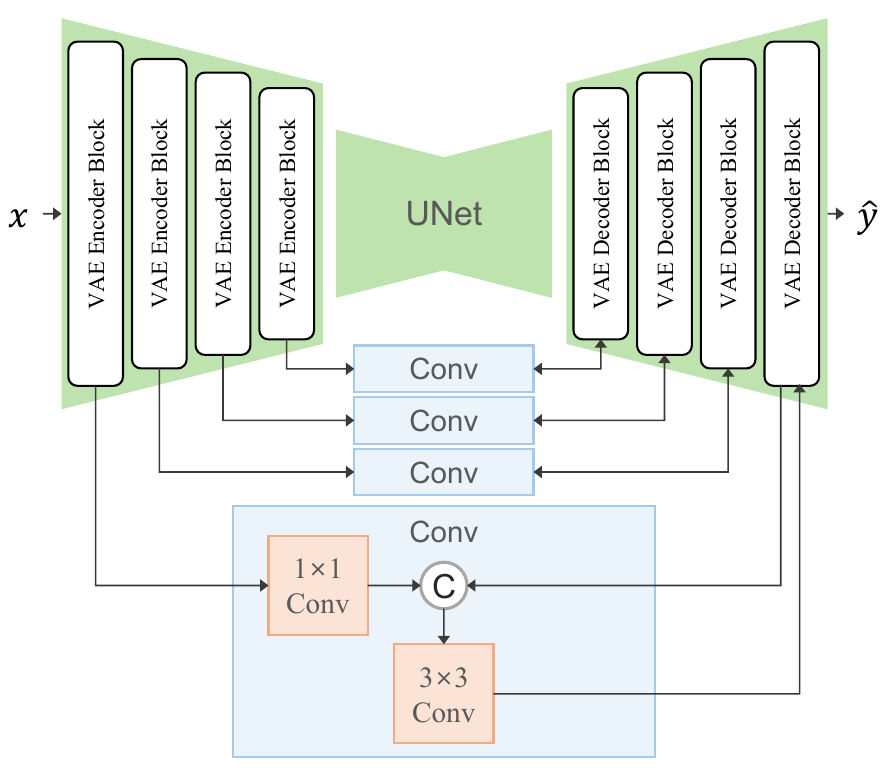}
    \caption{Detailed illustration of VAE skip-connection.}
    \label{fig:suppl_vae_structure}
\end{figure}

As described in Sec.~\ref{sec:odd}, we introduce lightweight two-layer convolutional skip connections (ODD-Conv) at each encoder-decoder level of the VAE to mitigate the detail loss commonly observed in the original Stable Diffusion VAE.
Specifically, as shown in Fig.~\ref{fig:suppl_vae_structure}, we extract intermediate encoder features and pass them through a $1\times1$ convolution to match the channel dimensions of the corresponding decoder features.
The transformed encoder features are then concatenated with the decoder features, followed by a $3\times3$ convolution to fuse the information effectively.
This design preserves fine-grained textures and structural details that would otherwise be lost during the VAE encoding-decoding process.

\begin{table}[t]
\centering
\caption{Ablation of ODD-Conv on AISTD. ``w/o Conv'' removes the VAE skip connections and is retrained from scratch.}
\label{tab:odd_conv}
\resizebox{0.65\linewidth}{!}{
\begin{tabular}{l|c|c|cc}
\toprule
\rowcolor{blue!7}
 & \textbf{Shadow} & \textbf{Non-Shadow} & \multicolumn{2}{c}{\textbf{ALL}} \\
\rowcolor{blue!7}
Method & PSNR$\uparrow$ & PSNR$\uparrow$ & PSNR$\uparrow$ & RMSE$\downarrow$ \\
\midrule
ODD w/o Conv & 33.14 & 27.79 & 26.16 & 8.46 \\
\rowcolor{gray!10}
ODD          & \best{36.67} & \best{37.45} & \best{33.32} & \best{3.11} \\
\bottomrule
\end{tabular}}
\end{table}

\noindent\textbf{Ablation.}
As shown in Tab.~\ref{tab:odd_conv}, removing ODD-Conv sharply degrades quality: ALL PSNR drops from 33.32 to 26.16\,dB, and the non-shadow region suffers the largest loss ($-$9.66\,dB).
Without the skip connections, details cannot bypass the lossy latent bottleneck, so even regions that should remain unchanged are corrupted.

\section{Transferability of ShadowReward}
\label{supp:transfer}

\begin{table}[t]
\centering
\caption{Transfer to ambient lighting normalization on Ambient6K~\cite{vasluianu2024ambient}. ODDR refines ODD on Ambient6K images using only the unmodified ShadowReward ($\lambda_r{=}0.1$), without ground truth.}
\label{tab:ambient6k}
\resizebox{0.5\linewidth}{!}{
\begin{tabular}{l|cc}
\toprule
\rowcolor{blue!7}
Method & PSNR$\uparrow$ & SSIM$\uparrow$ \\
\midrule
DeS3~\cite{jin2024des3} & 14.44 & 0.681 \\
SSR~\cite{xu2025detail} & 15.64 & 0.704 \\
ODD (zero-shot) & 15.65 & 0.719 \\
\rowcolor{gray!10}
ODDR (reward-guided, no GT) & \best{17.16} & \best{0.747} \\
\midrule
$\Delta$ (reward) & +1.51 & +0.028 \\
\bottomrule
\end{tabular}}
\end{table}

\subsection{Ambient Lighting Normalization}
Our main experiments focus on cast-shadow benchmarks.
We further examine ambient lighting normalization, where the degradation is a global, smooth, color-dominated illumination field rather than a discrete masked shadow, using the Ambient6K benchmark~\cite{vasluianu2024ambient}.
Following the same recipe as for shadows, ODD serves as the base model and the ODDR stage refines it on the official training images of Ambient6K using only the unmodified ShadowReward ($\lambda_r{=}0.1$), with no ground truth; evaluation is performed on the official test split.
As shown in Tab.~\ref{tab:ambient6k}, the reward stage improves the base model by +1.51\,dB PSNR without any paired supervision, and ODDR also scores higher than the diffusion-based shadow removal methods DeS3 and SSR.
We report this as an ODD$\rightarrow$ODDR gain rather than a comparison with dedicated ambient lighting methods.
The transfer is plausible because two of the five degradation types, chromatic shift and luminance change, model the same kind of color and brightness change that ambient lighting causes.
Moreover, the shadow matte used by SynShadow only shapes how the training data of ShadowReward is constructed; at inference, both ODDR and ShadowReward operate without masks.
The low absolute score of the AISTD-trained base, however, shows that complex, physically grounded lighting is not covered by our synthetic shadow pipeline (see Sec.~\ref{supp:limitations}).
Recent work on generalized shadow removal with physically aligned priors~\cite{lee2026phasr} is complementary to our annotation-free approach.

\begin{table}[t]
\centering
\caption{Transfer of ShadowReward to other low-level vision tasks. PromptIR~\cite{potlapalli2023promptir}, trained on synthetic data, is refined with the unmodified ShadowReward on real degraded images only (no ground truth). PSNR / SSIM.}
\label{tab:cross_task}
\resizebox{0.7\linewidth}{!}{
\begin{tabular}{l|c|c}
\toprule
\rowcolor{blue!7}
Real-world benchmark & Base & +ShadowReward \\
\midrule
Denoising (SIDD~\cite{abdelhamed2018sidd}) & 22.84 / 0.341 & \best{23.35} / \best{0.344} \\
Deraining (GT-RAIN~\cite{ba2022gtrain}) & 21.28 / 0.632 & \best{22.06} / \best{0.636} \\
Dehazing (NH-HAZE~\cite{ancuti2020nhhaze}) & 12.33 / 0.444 & 12.32 / 0.444 \\
\bottomrule
\end{tabular}}
\end{table}

\subsection{Other Low-Level Vision Tasks}
The recipe behind ShadowReward is not specific to shadows: it requires only clean images and a taxonomy of degradations that reproduce the artifacts a restoration model tends to leave behind.
To examine whether the learned reward itself transfers, we take PromptIR~\cite{potlapalli2023promptir}, an all-in-one restorer trained on synthetic data, and refine it with the unmodified ShadowReward on real-world benchmarks for three tasks: SIDD~\cite{abdelhamed2018sidd} (real sensor noise), GT-RAIN~\cite{ba2022gtrain} (real rain), and NH-HAZE~\cite{ancuti2020nhhaze} (real non-homogeneous haze).
We follow the official data splits of each benchmark, using only the degraded images for refinement and the ground truth solely for evaluation.
Since the base model has never seen these real domains, the absolute scores are low; what matters is the gain from the reward.
As shown in Tab.~\ref{tab:cross_task}, the gain follows the coverage of ShadowReward's degradation types.
Additive noise is one of the five types, so the reward improves denoising.
Deraining, like deshadowing, requires recovering specific local content, and the reward likewise transfers.
In contrast, ShadowReward never observes haze-type degradation and thus provides no signal for dehazing.
Extending the degradation taxonomy with physically grounded haze modeling, as explored in recent dehazing work~\cite{im2023vdhr, shin2025hazeflow}, is a natural direction for future work.
More broadly, since reward-guided refinement only requires a one-step restorer and a task-aware reward, it could similarly be applied to one-step models for other restoration tasks, such as dehazing~\cite{shin2025hazeflow}, reflection removal~\cite{shin2026prism}, and deblurring~\cite{shin2026pickmoment}.

\section{Limitations and Failure Cases}
\label{supp:limitations}

\begin{figure}
    \centering
    \includegraphics[width=0.85\linewidth]{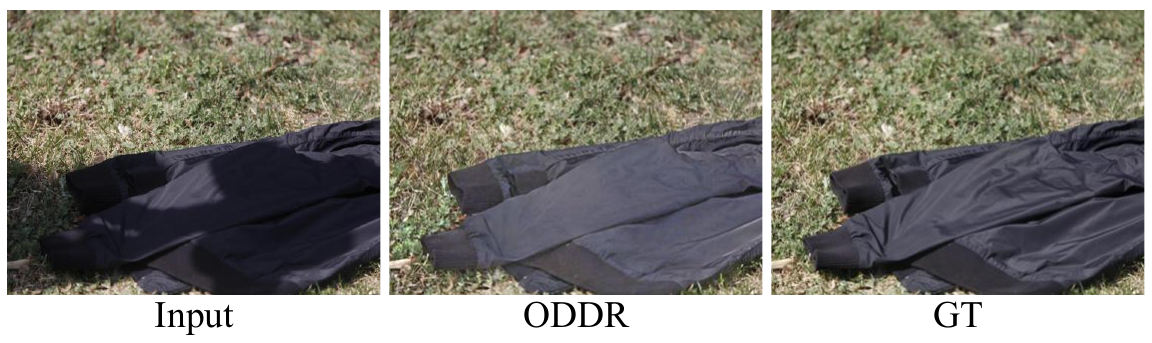}
    \caption{Example of failure case on highly textured/non-flat surfaces.}
    \label{fig:suppl_limitations}
\end{figure}

While ODDR achieves strong shadow removal performance without real-world paired supervision, several limitations remain.
First, on extremely hard shadow boundaries with large intensity gaps, ODDR can leave faint halos or over-smooth the transition, since ShadowReward's boundary degradation (B) is trained with only mild dilation and Gaussian blur and thus does not fully cover such out-of-distribution edges.
Second, on highly textured or non-flat surfaces (e.g., grass, foliage, or fabric), the recovered shadow region may exhibit inconsistent texture or subtle color drift, as synthetic degradations do not faithfully reproduce these surface-specific interactions (Fig.~\ref{fig:suppl_limitations}).
Third, ShadowReward is trained on five synthetic degradation types (texture distortion, additive noise, chromatic shift, luminance change, and boundary artifact), and therefore provides limited supervision against artifacts that lie outside this taxonomy, such as semantic inconsistency or geometric misalignment introduced by the diffusion prior, or haze-type degradation, for which it provides no signal (Tab.~\ref{tab:cross_task}).
Fourth, since ODD is trained with synthetic shadows generated from AISTD, it handles complex scenes less well, including penumbra-dominated, multiple or overlapping, and self-shadows; on the physically grounded Ambient6K benchmark, the base model reaches only 15.65\,dB PSNR (Tab.~\ref{tab:ambient6k}). Although ShadowReward still improves it, diversifying the synthetic shadow pipeline to cover such cases remains future work.
Finally, our framework is developed under the standard single-image setting with a static scene; extensions to video shadow removal, multi-light scenes, and joint reasoning with scene geometry are promising directions for future work.

\noindent\textbf{Failure cases of ShadowReward.}
Regarding the 5 of 16 user-study questions where ShadowReward disagrees with the human consensus, its ranking is least reliable on very faint shadows, where the residual shadow lies near the noise floor and the candidates differ only marginally; such cases are also difficult for human annotators to rank consistently.
The remaining disagreements involve artifacts outside the five degradation types, such as diffusion-induced texture or structure changes, which we regard as the current coverage boundary of the taxonomy.

\paragraph{Broader impacts.}
ODDR is a low-level restoration method with limited misuse risk; like any image-editing tool, restored outputs should be treated as edited rather than original.

\section{Licenses of Existing Assets}
\label{supp:licenses}

The proposed framework builds on the following publicly available assets, which we use within the bounds of their respective licenses:
\begin{itemize}
    \item \textbf{Stable Diffusion v2.1}~\cite{rombach2022stablediffusion}~--~CreativeML Open RAIL++-M License.
    \item \textbf{DINOv2}~\cite{oquab2023dinov2}~--~Apache License 2.0.
    \item \textbf{OSEDiff}~\cite{wu2024osediff}~--~Apache License 2.0.
    \item \textbf{LoRA}~\cite{hu2022lora}~--~MIT License.
    \item \textbf{LPIPS}~\cite{zhang2018unreasonable}~--~BSD 2-Clause License.
    \item \textbf{SynShadow}~\cite{inoue2020learning}~--~MIT License.
    \item \textbf{PromptIR}~\cite{potlapalli2023promptir}~--~Academic Public License (non-commercial research use).
    \item \textbf{AISTD/ISTD+}~\cite{wang2018istd, le2019shadow}, \textbf{SRD}~\cite{qu2017srd}, \textbf{LRSS}~\cite{gryka2015learninglrss}, \textbf{UIUC}~\cite{guo2012paireduiuc}, \textbf{Ambient6K}~\cite{vasluianu2024ambient}, \textbf{SIDD}~\cite{abdelhamed2018sidd}, \textbf{GT-RAIN}~\cite{ba2022gtrain}, and \textbf{NH-HAZE}~\cite{ancuti2020nhhaze}~--~released for academic / research use as specified by the original authors.
\end{itemize}

\section{Additional Visual Results}
\label{supp:add_visual_results}

\begin{figure*}
    \centering
    \includegraphics[width=\linewidth, height=0.95\textheight, keepaspectratio]{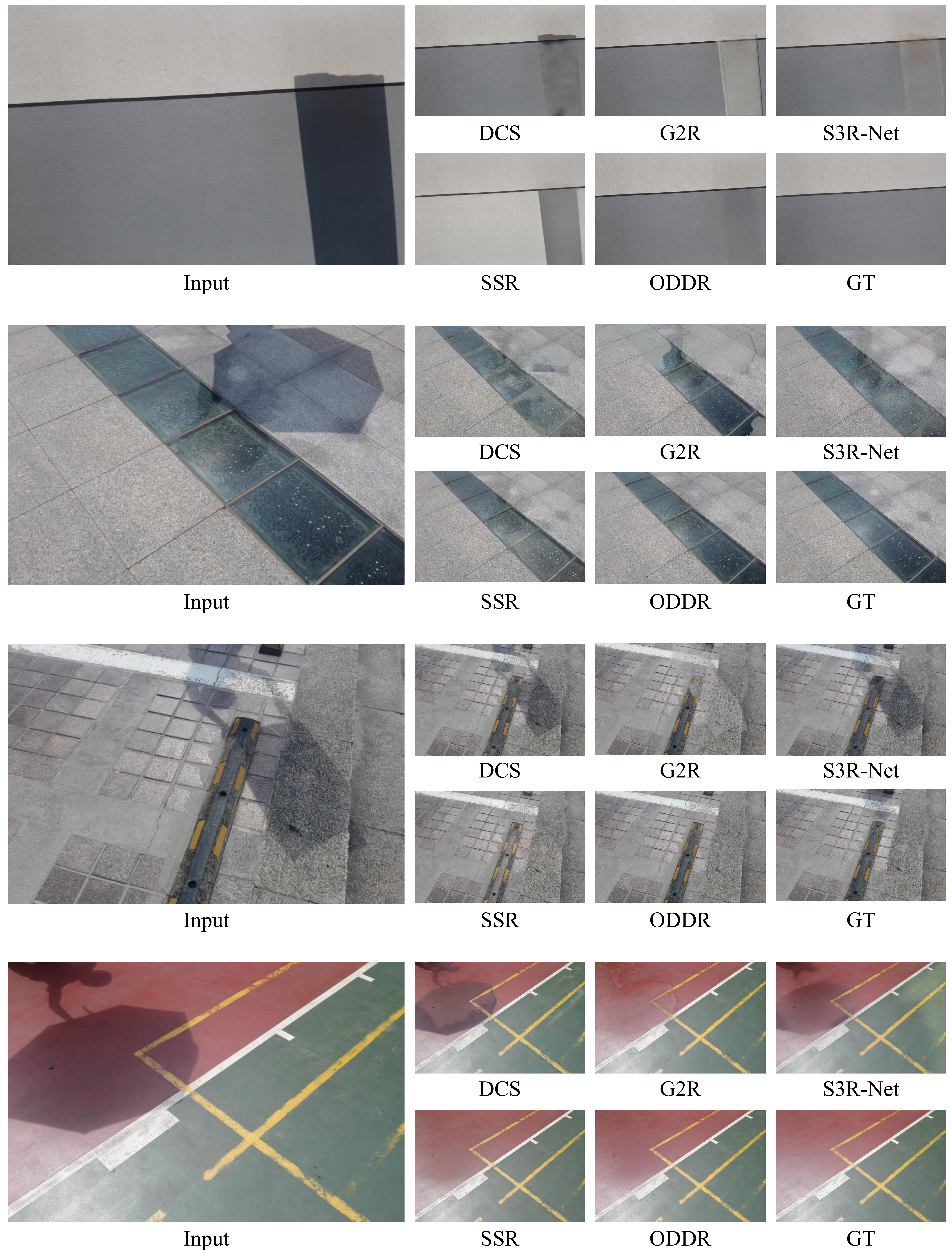}
    \caption{Additional visual comparison on the AISTD~\cite{le2019shadow} dataset.}
    \label{fig:suppl_aistd}
\end{figure*}

\begin{figure*}
    \centering
    \includegraphics[width=\linewidth]{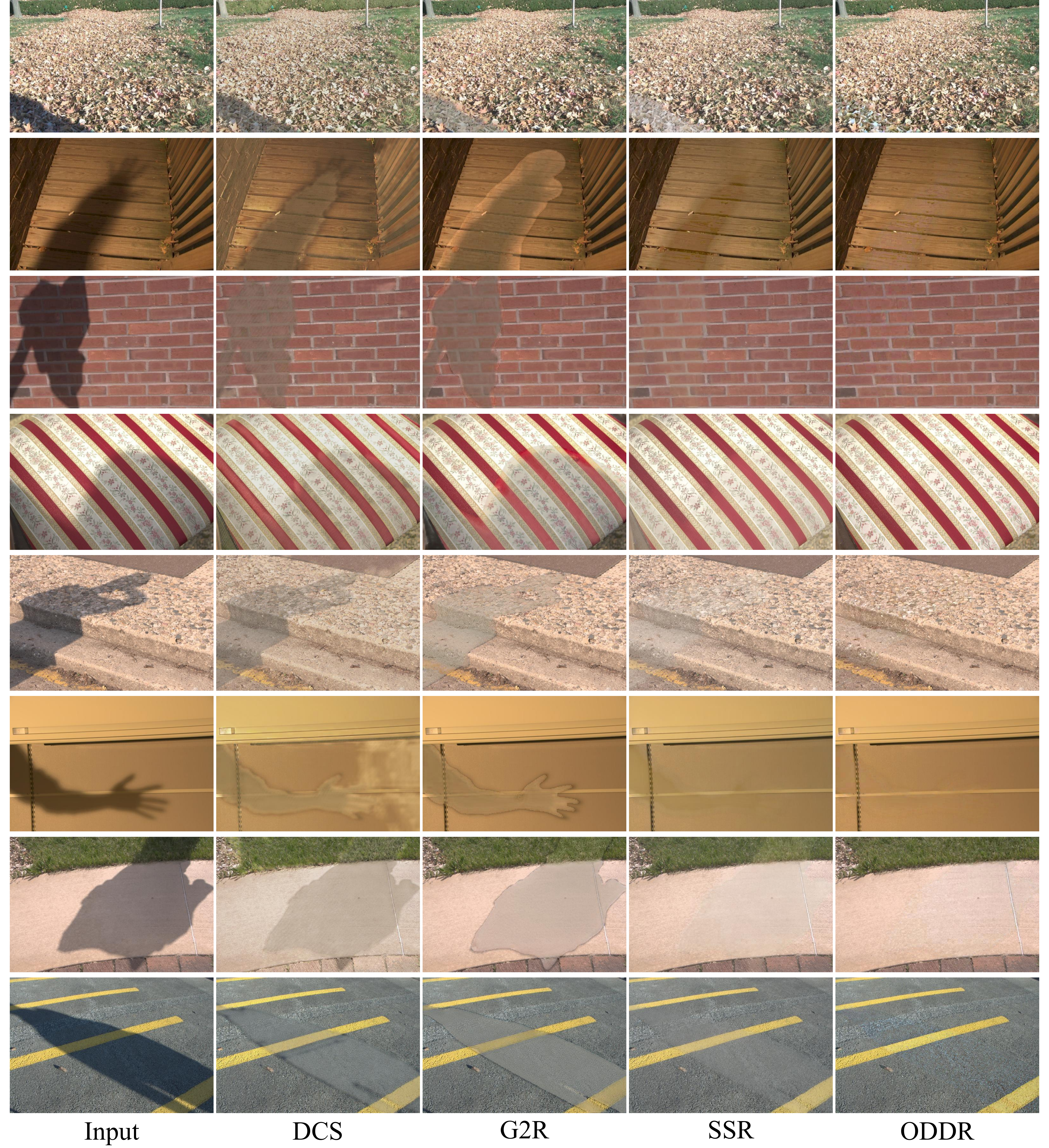}
    \caption{Additional visual comparisons on the LRSS~\cite{gryka2015learninglrss} (top four rows) and UIUC~\cite{guo2012paireduiuc} (bottom four rows) datasets.}
    \label{fig:suppl_lrss_uiuc}
\end{figure*}

In this section, we provide additional visual results to further demonstrate the effectiveness and robustness of our proposed method across diverse scenes and conditions.
As shown in Fig.~\ref{fig:suppl_aistd}, our method achieves significantly better visual quality on the AISTD~\cite{le2019shadow} dataset than existing unsupervised shadow removal approaches, producing cleaner shadow-free regions and more consistent illumination.
Moreover, even when compared with supervised methods, our results exhibit only marginal differences, highlighting the strong generalization capability of our framework despite the absence of paired supervision.
Fig.~\ref{fig:suppl_lrss_uiuc} presents qualitative results on unseen datasets~\cite{gryka2015learninglrss, guo2012paireduiuc}, demonstrating the strong generalization ability of our ODDR framework. 
Despite never being trained on these datasets, ODDR consistently removes shadows while preserving natural color and structure, whereas existing methods often leave residual shadows or introduce color distortions. These results further confirm that ODDR is highly robust to domain shifts and remains effective even in challenging unseen scenarios.

\begin{figure*}
    \centering
    \includegraphics[width=\linewidth]{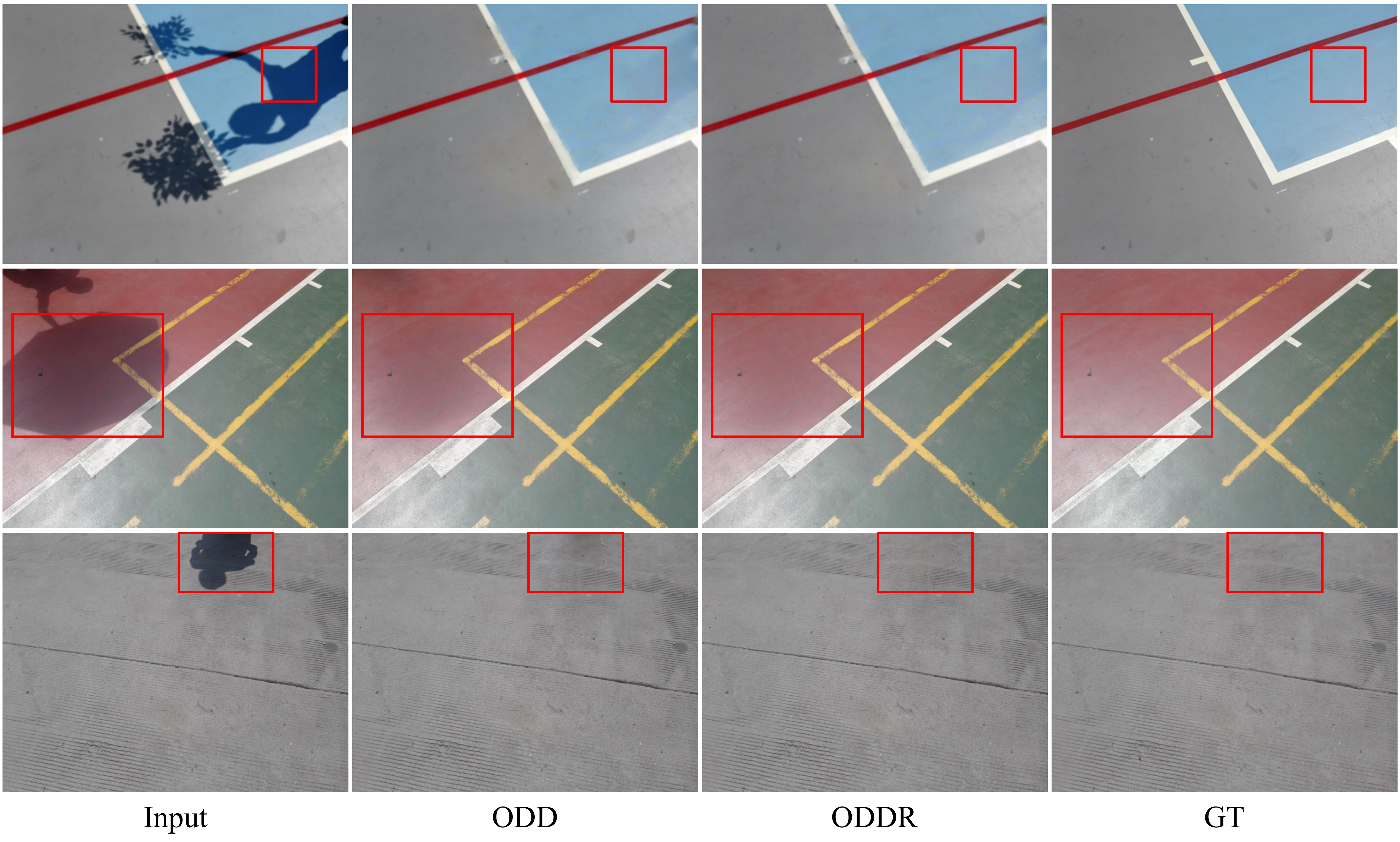}
    \caption{Additional visual comparison of ODD and ODDR on the AISTD~\cite{le2019shadow} dataset.}
    \label{fig:suppl_odd_oddr_aistd}
\end{figure*}

\begin{figure*}
    \centering
    \includegraphics[width=\linewidth]{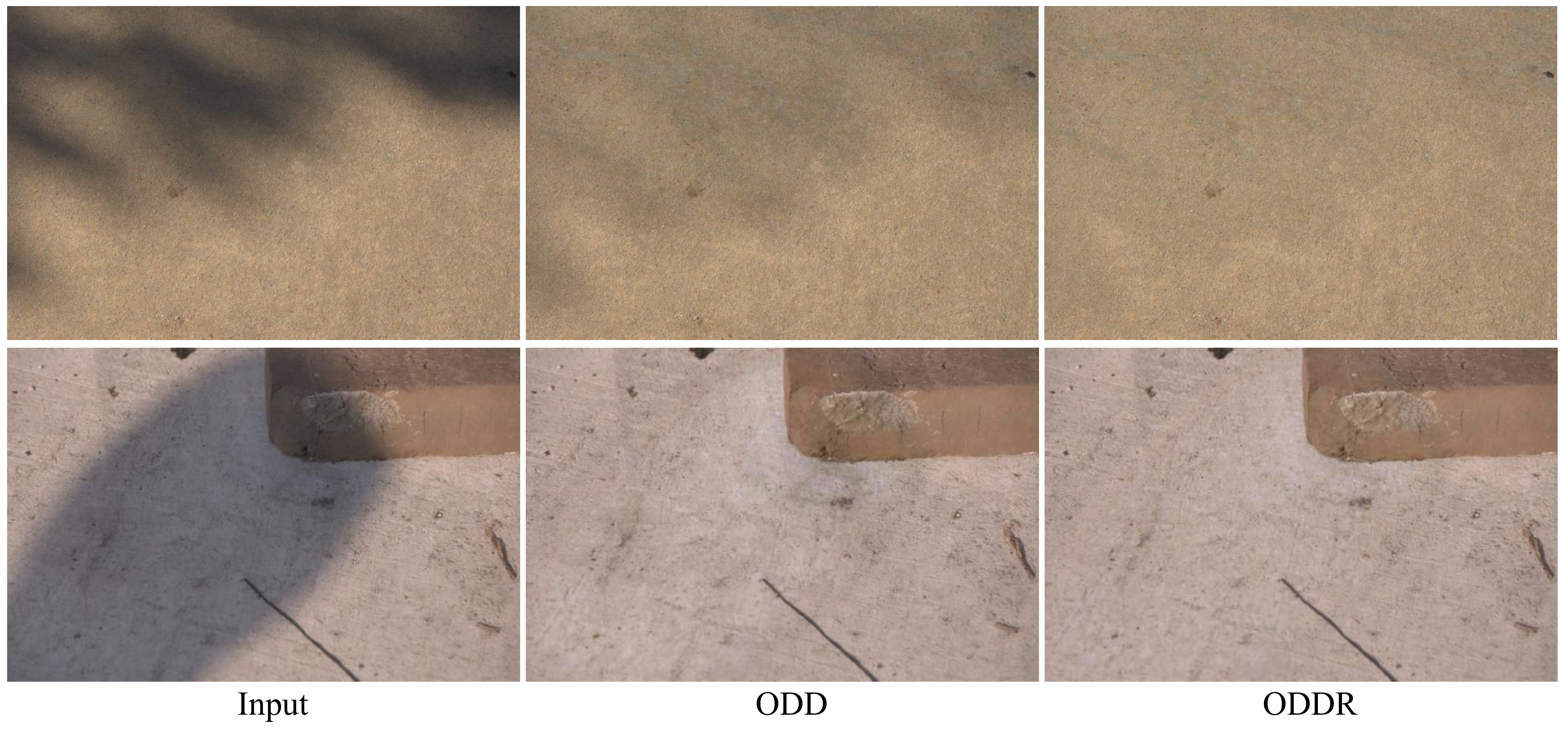}
    \caption{Additional visual comparison of ODD and ODDR on the unseen LRSS~\cite{gryka2015learninglrss} dataset.}
    \label{fig:suppl_odd_oddr_lrss}
\end{figure*}

\begin{figure*}
    \centering
    \includegraphics[width=\linewidth]{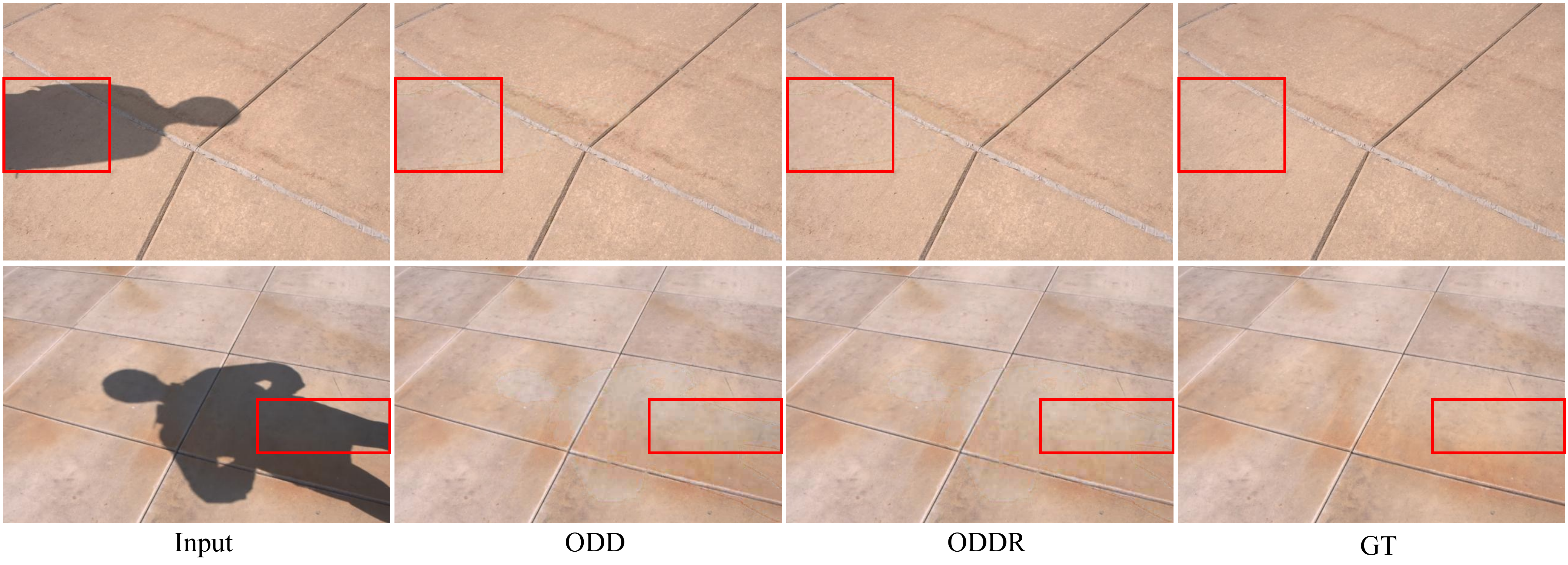}
    \caption{Additional visual comparison of ODD and ODDR on the unseen UIUC~\cite{guo2012paireduiuc} dataset.}
    \label{fig:suppl_odd_oddr_uiuc}
\end{figure*}

To provide a more comprehensive analysis of our framework, we present additional visual comparisons between ODD and ODDR across multiple datasets. 
Fig.~\ref{fig:suppl_odd_oddr_aistd} shows qualitative results on the AISTD dataset, where ODDR produces cleaner shadow removal and more stable color consistency than ODD. 
Furthermore, as illustrated in Fig.~\ref{fig:suppl_odd_oddr_lrss} and Fig.~\ref{fig:suppl_odd_oddr_uiuc}, ODDR demonstrates superior performance on unseen datasets such as LRSS and UIUC, effectively preserving structural details while removing shadows that ODD occasionally fails to eliminate. These results collectively highlight the advantages of reward-guided optimization and confirm that ODDR delivers more robust and perceptually consistent restoration across both seen and unseen domains.

